\documentclass[10pt,twocolumn,letterpaper]{article}

\usepackage[pagenumbers]{cvpr}

\newif\ifarxivversion
\arxivversionfalse

\usepackage{placeins}
\usepackage{subcaption}

\arxivversiontrue

\definecolor{cvprblue}{rgb}{0.21,0.49,0.74}
\usepackage[pagebackref,breaklinks,colorlinks,allcolors=cvprblue]{hyperref}
\usepackage{comment}
\usepackage{bibunits}
\usepackage{etoolbox}
\makeatletter
\patchcmd{\@startbibunit}{\endlinechar-1}{\endlinechar-1\makeatletter}{}{\PackageWarning{main_arxiv}{bibunits patch failed}}
\patchcmd{\@finishstartedbibunit}{\endlinechar-1}{\endlinechar-1\makeatletter}{}{\PackageWarning{main_arxiv}{bibunits patch failed}}
\makeatother

\def\confName{CVPR}
\def\confYear{2027}

\newcommand{\vlt}{VLT\xspace}
\newcommand{\vltgenname}{ThinkerBell}
\newcommand{\vltgen}{\vltgenname\xspace}
\definecolor{thinkerBellGreen}{HTML}{527A1D}
\newcommand{\onlygen}{generator-only\xspace}

\newcommand{\sig}{$^{\star}$}

\newcommand{\authorcomment}[3]{{\color{#1}\textbf{#2$\rightarrow$}~#3}}
\definecolor{mjMutedMagenta}{HTML}{A0457E}
\definecolor{wwPaleBlue}{HTML}{6F9FD8}
\ifodd 0
  \newcommand{\mj}[1]{\authorcomment{mjMutedMagenta}{mj}{#1}}
  \newcommand{\sy}[1]{\authorcomment{ForestGreen}{sy}{#1}}
  \newcommand{\yj}[1]{\authorcomment{BurntOrange}{yj}{#1}}
  \newcommand{\ds}[1]{\authorcomment{Plum}{ds}{#1}}
  \newcommand{\ww}[1]{\authorcomment{wwPaleBlue}{ww}{#1}}
  \newcommand{\tk}[1]{\authorcomment{TealBlue}{tk}{#1}}
\else
  \newcommand{\mj}[1]{}
  \newcommand{\sy}[1]{}
  \newcommand{\yj}[1]{}
  \newcommand{\ds}[1]{}
  \newcommand{\ww}[1]{}
  \newcommand{\tk}[1]{}
\fi

\makeatletter
\g@addto@macro{\@maketitle}{\begin{center}
  \includegraphics[width=\textwidth]{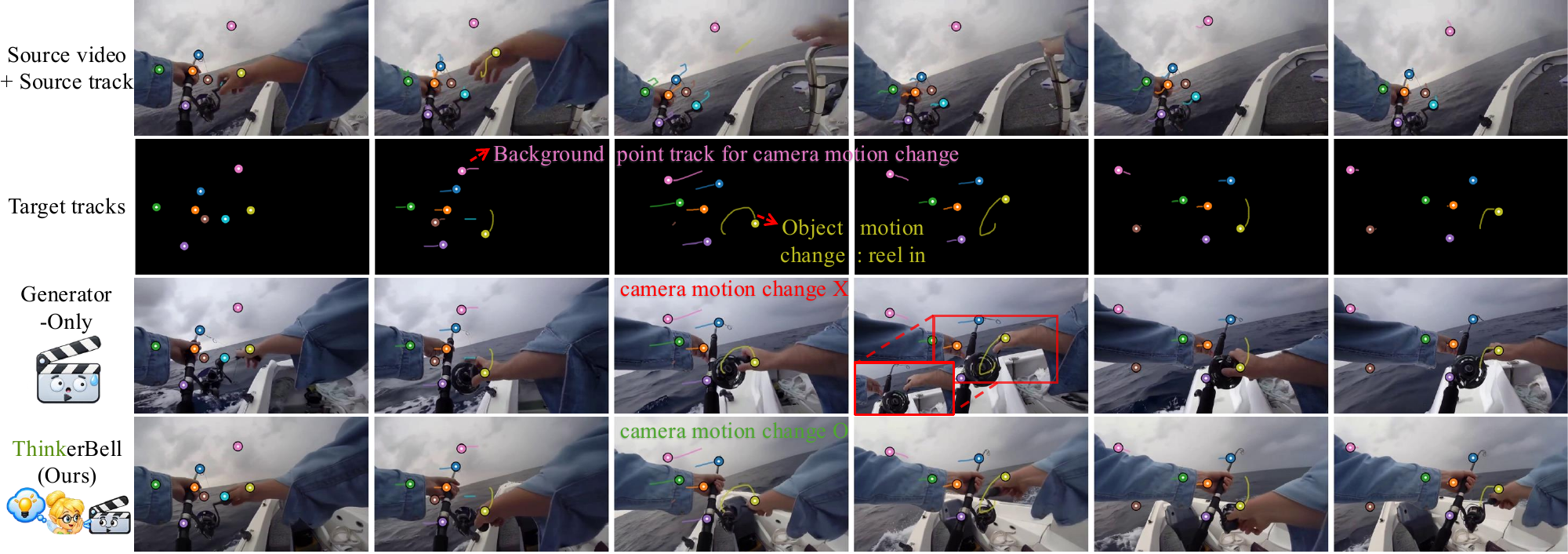}
  \vspace{1pt}
  \begin{minipage}{\textwidth}\small\refstepcounter{figure}\label{fig:teaser}\textbf{Figure~\thefigure.}~\textbf{Track-conditioned video motion editing with \vltgen.}
   The first two rows show the source video with source tracks and the target tracks that condition generation (captions and instructions omitted).
   In an egocentric fishing video, the target tracks bring the free hand to the reel to crank it and rotate the camera slightly to the right.
   A generator-only baseline~\cite{motionv2v} infers this motion implicitly from the sparse tracks, misses the camera motion, and breaks the reel and hand (red boxes).
   \vltgen first interprets which objects the tracks describe and how they and the camera should move, then follows both while preserving hands, rod, and boat.
   \ifarxivversion\else
   \emph{We kindly invite the reader to view the videos in the supplementary HTML.}
   \fi
  \end{minipage}
\end{center}
}
\makeatother

\newcommand{\titleemoji}{%
  \makebox[0pt][r]{\raisebox{-5pt}[17pt][5pt]{%
    \includegraphics[height=44pt]{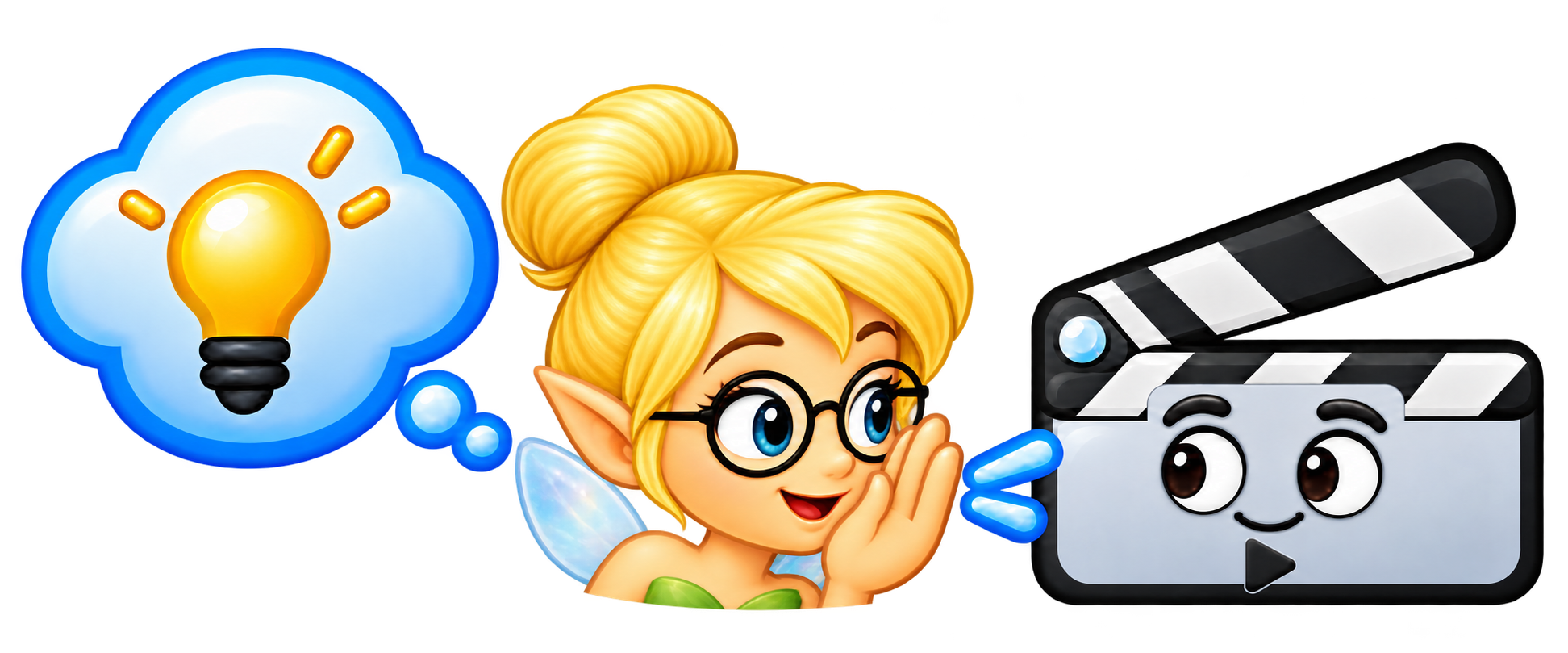}}\hspace{0.1em}}%
}
\title{\texorpdfstring{\hspace*{8pt}\titleemoji}{}%
\texorpdfstring{\textcolor{thinkerBellGreen}{Think}erBell}{\vltgenname}: Understanding Before Generation\\
for Track-Conditioned Video Motion Editing}

\author{%
Minjung Shin$^{1*}$ \quad
Sooyeon Go$^{1}$ \quad
Dongseok Shim$^{2}$ \quad
Takashi Shibuya$^{2}$\\[3pt]
Shusuke Takahashi$^{2}$ \quad
Wei-Yao Wang$^{2\dagger}$ \quad
Youngjung Uh$^{1\dagger}$ \quad
Yuki Mitsufuji$^{2\dagger}$\\[6pt]
$^{1}$Yonsei University \qquad $^{2}$Sony Group Corporation\\[6pt]
{\normalsize\textbf{Project Page : \href{https://minjung-s.github.io/thinkerbell}{\texttt{https://minjung-s.github.io/thinkerbell}}}}
}

\begin{document}
\addtocontents{toc}{\protect\setcounter{tocdepth}{-1}}
\maketitle
\begingroup
\renewcommand{\thefootnote}{\fnsymbol{footnote}}
\footnotetext[1]{Part of this work was done during an internship at Sony.}
\footnotetext[2]{Co-corresponding authors.}
\endgroup
\begin{abstract}

Video motion editing aims to change object or camera motion in a source video while preserving its content.
Sparse point trajectories are a convenient way to specify the target motion, yet existing methods treat them only as geometric constraints and leave the generator to infer what motion they imply, which fails under complex edits.
We propose to \textit{understand the sparse tracks before generation}.
\textit{\textcolor{thinkerBellGreen}{Think}erBell} couples a vision--language--track reasoner (VLT) to a video generator.
We propose \vlt, trained on 1.9M track-understanding questions: which scene objects the tracks follow, how the objects and the camera move, and whether the source--target track pairs agree with the given editing instruction.
To deliver this understanding to a generator, we design a dedicated learnable coupling that transfers the hidden states of \vlt to the generator across their different depths, widths, and token representations.
We further introduce a benchmark of real source--target video pairs that separates motion types into object, camera, and joint object--camera motion editing across diverse scenes.
\vltgen sets a new state of the art in target-video fidelity in every category and in trajectory controllability in all categories that involve camera motion.
Analyses show that stronger track understanding yields better generation, and that the gains hold across generators with different trajectory-conditioning mechanisms.

\end{abstract}

\section{Introduction}
\label{sec:intro}

Video motion editing aims to modify object or camera motion in a source video while preserving object appearance and background content. 
Typically, users provide a source video, source point tracks following the source motion, target point tracks, and a target description~\cite{motionv2v,editbytrack}.
In track-conditioned video-to-video (V2V) editing, point tracks provide an intuitive interface for controlling object motion, camera motion, or both through target trajectories that prescribe new paths for points tracked in the source video.
Track-conditioned image-to-video (I2V) generation has advanced rapidly~\cite{flextraj,ati,tora,wanmove}, but it starts from a single reference frame.
Editing a video requires more: objects that appear or become occluded in later frames are visible only across the source frames, which only V2V editing can preserve.
For applications such as video post-production and visual effects, specifying motion with a few sparse 2D points avoids the labor-intensive process of defining dense or 3D target trajectories~\cite{editbytrack,unicamo}.

However, existing V2V methods conditioned on sparse 2D tracks often introduce visual artifacts or fail to preserve source content under complex motion changes~\cite{motionv2v} (\cref{fig:teaser}).
We attribute these failures to the gap between what a few sparse tracks specify and the full motion of objects, the camera, and the scene that they are meant to control.
Existing approaches~\cite{motionv2v,revideo} focus on incorporating sparse tracks into the generation process, leaving the generator to infer the underlying motion semantics implicitly through the denoising objective alone.
Because sparse tracks provide limited information about how the intended motion should interact with scene geometry, object dynamics, and source content, this implicit reasoning is particularly challenging under complex edits.

To bridge this gap, we propose a new perspective: \textit{explicitly understand the sparse tracks before generation}.
Our key intuition is that this understanding should come from a separate reasoner trained to interpret the tracks, whose representations then tell the generator which motion the sparse constraints intend and which scene content to preserve.
We therefore introduce \textit{\textcolor{thinkerBellGreen}{Think}erBell}, a framework in which a vision--language--track reasoner (VLT) understands the editing intent specified by the tracks and a video generator receives this understanding and only synthesizes the edited video.
\vlt interprets the requested edit, and its representations guide the generator alongside direct conditioning on the given target tracks.

Existing vision--language models (VLMs) fall short as this reasoner: without interpreting the given target tracks in relation to the source scene and the editing intent, they can merely output coordinates.
Our reasoner \vlt addresses this limitation by jointly interpreting the source video, source tracks, target tracks, and an editing instruction that describes the requested change to the source motion.
We train \vlt on track-understanding tasks, using a curriculum of 1.9M questions that ties track geometry to scene objects and actions, then relates changes in their motion to editing instructions.
Through this training, \vlt learns representations that connect the tracks to the source scene and to the editing instruction: which object or camera the tracks follow, what motion they express, and whether the change from source to target tracks carries out the requested edit.
This understanding, rather than the coordinates alone, is what \vlt passes to the generator as hidden states.

Because \vlt and the generator are pretrained independently with different architectures and training objectives, we develop a coupling that accommodates their differences in depth, width and token representations.
We design this coupling around three choices:
(i)~which tokens to transfer, by selecting the edit-relevant token spans;
(ii)~where to connect the two models, by learning which reasoner layer feeds each generator block while retaining both models' original depths;
and (iii)~how to project the transferred tokens, with a projection specific to each token type.

To test whether this understanding improves motion editing, we propose a benchmark for track-conditioned video motion editing with four categories: joint object--camera, object-only, and camera-only motion editing in dynamic and static scenes.
Each pair consists of two real videos of the same scene that differ in motion, with corresponding point tracks.
We measure track error not only at the conditioned tracks but also at unconditioned ones, which tests whether the whole object or region indicated by the sparse tracks moves with them.
\vltgen sets a new state of the art in target-video fidelity in all four categories and in track error in the three categories that involve camera motion.
Our analyses further show that the track-understanding tasks give \vlt its ability to interpret tracks in relation to editing instructions, that this understanding translates into better video generation, and that the benefit holds across generators with different track-conditioning mechanisms.

Our contributions are as follows:
\begin{itemize}[leftmargin=*]
  \item We introduce \vltgen, a track-conditioned video motion editing framework that integrates explicit motion understanding with trajectory-guided generation.
  \item We develop \vlt, a vision--language--track reasoner that jointly interprets source videos, source and target trajectories, and editing instructions to learn scene-grounded representations of the intended motion changes.
  \item We propose a learnable coupling that transfers edit-relevant reasoner representations to independently pretrained generators of different depth, width and token representations.
  \item We propose a benchmark for track-conditioned video motion editing that covers object-only, camera-only and joint object--camera edits. \vltgen achieves the best target-video fidelity in every benchmark category and the lowest track error whenever the edit involves camera motion.
\end{itemize}

\section{Related Work}
\label{sec:related}

\paragraph{Track-conditioned video editing.}
Track-conditioned I2V approaches animate a reference image along specified trajectories~\cite{dragnuwa,draganything,motioni2v,imageconductor,tora,motionprompting,ati,flextraj,wanmove}, but inherently lack access to temporal information beyond the reference frame.
V2V methods instead use additional source frames to preserve scene coherence~\cite{editbytrack,motionv2v,gowiththetrack}; Edit-by-Track~\cite{editbytrack} edits complex camera and object motion through paired dense 3D tracks, but specifying dense target trajectories is time-consuming, which motivates sparse 2D tracks as a more practical interface.
Methods using sparse 2D tracks differ in which motions they can change and which parts of the source video they use as input.
TrajAttn~\cite{trajattn} changes camera motion while retaining source object dynamics, but does not directly support changing a selected object's motion.
ReVideo~\cite{revideo} edits motion within a user-specified region, but it masks that region out of the source video and regenerates its content from the edited first frame, so source details that appear only in later frames are lost.
Other approaches use additional source frames more directly: Go-with-the-Track incorporates multiple reference frames~\cite{gowiththetrack}, and MotionV2V conditions on the source video to edit motion using sparse 2D tracks~\cite{motionv2v}.
These methods still produce artifacts and degrade source content under complex edits (\cref{fig:teaser}), which we attribute to leaving track interpretation to the generator; we therefore add a dedicated reasoner that interprets the tracks with the source scene and the editing intent before generation.

\paragraph{Connecting understanding to generation.}
Multimodal understanding models can supply a generator with what the denoising objective does not explicitly supervise.
One line of work passes explicit outputs: descriptions or plans written by language models~\cite{freecus,rpg,planx}, or the trajectories and prompts that MotiMotion~\cite{motimotion} refines with a VLM for track-conditioned I2V generation.
Only the decoded content reaches the generator, discarding the understanding behind it, and decoding adds inference cost.
Another line transfers internal features.
InstructX~\cite{instructx} and MetaQueries~\cite{metaquery} read the VLM through learnable queries that condition the generator.
BAGEL and Cosmos 3 let each generator layer attend to the corresponding layer of an understanding tower~\cite{bagel,cosmos3}, which requires matched depths.
UniFusion~\cite{unifusion} relaxes this by pooling features across the layers of a frozen VLM for image generation, and MoRoute~\cite{moroute} by routing each block of a video generator to one VLM layer selected from the input.
None of these reasoners is trained to interpret point tracks together with the source scene and a motion-editing instruction.
To fill this gap, we train \vlt for this joint interpretation (\cref{sec:vlt}) and transfer its internal prompt representations to a generator of different depth, width and token content without decoding an answer.

\section{\vltgenname}
\label{sec:method}
\begin{figure*}[t]
  \centering
  \includegraphics[width=\textwidth]{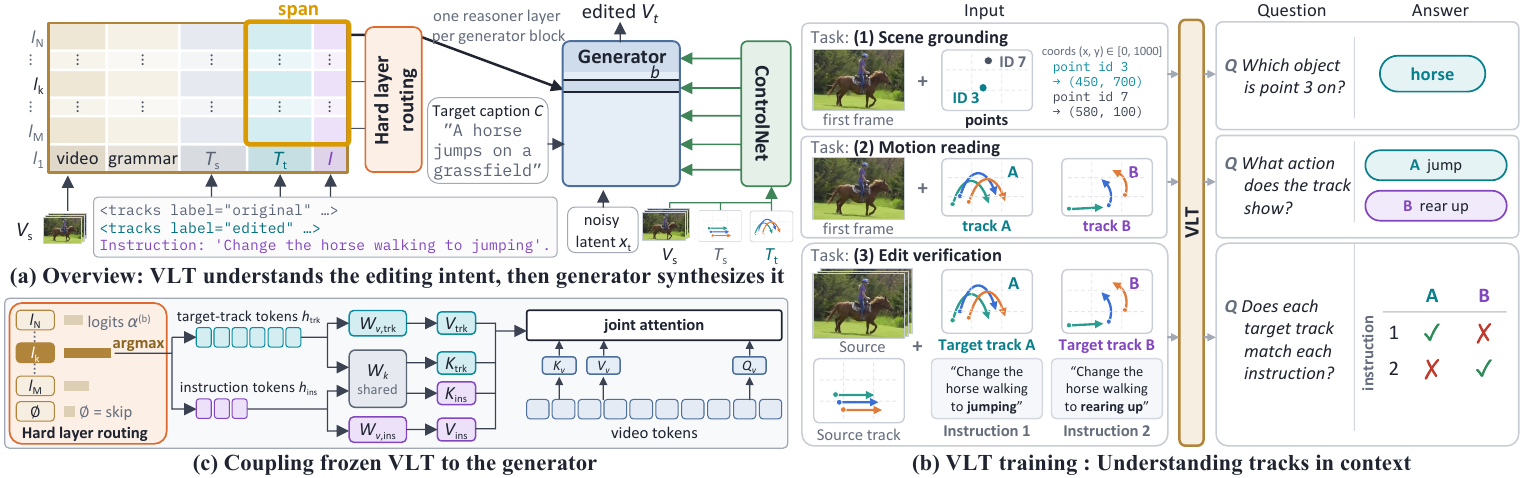}
  \captionsetup{skip=2pt}
  \caption{\textbf{\vltgen understands the editing intent before generating the video (a;~\cref{sec:overview}).}
  \vlt is trained to understand tracks in context (b;~\cref{sec:vlt}), then frozen and coupled to the generator via its prompt hidden states (c;~\cref{sec:vltgen}).
  }
  \label{fig:overview}
\end{figure*}

\subsection{Overview}
\label{sec:overview}
\paragraph{Setting.}
\vltgen addresses track-conditioned video motion editing: modifying object or camera motion in a video while preserving its content.
Given a source video $V_s$, the goal is to generate an edited video $V_t$ whose motion is specified by a user while preserving the source content.
The user selects a sparse set of source tracks $T_s$ extracted from $V_s$ and specifies their desired motion as corresponding target tracks $T_t$. Each source--target track pair follows the same physical point, aligned in time.
Two text inputs complete the specification: an editing instruction $I$ stating the change from the source motion (e.g., ``Change the horse walking to jumping''), and a target description $C$ captioning the edited video alone (e.g., ``A horse jumps on a grass field'').
The editing instruction is an input newly introduced by \vltgen and absent from existing track-conditioned methods; it is a short sentence that a user can write with little effort.
The task is challenging because the motion of a track can come from the object, the camera, or both, and sparse tracks leave the motion of untracked regions unspecified, which leads to distorted results when inferred incorrectly.

\paragraph{Approach.}
\vltgen assigns the task to two models: a reasoner, \vlt, that relates the tracks to the source scene and the editing instruction, and a video generator that synthesizes the edited video (\cref{fig:overview}a).
The generator takes $(V_s,T_s,T_t,C)$ as in existing track-conditioned generators, and additionally receives the hidden states (i.e., intermediate token features) that \vlt computes for $(V_s,T_s,T_t,I)$. These hidden states are expected to convey this relation, as \vlt is trained to answer questions about the tracks (\cref{sec:vlt}).
To bridge the architectural differences between the independently pretrained models, we introduce a learnable coupling that aligns and transfers the reasoner's hidden states to the generator across different network depths and feature dimensions (\cref{sec:vltgen}).

\subsection{\vlt: Understanding Tracks in Context}
\label{sec:vlt}
\label{sec:vlt_why}
We hypothesize that the generator benefits from receiving the hidden states of \vlt when they encode the relation of the tracks to the source scene and the editing instruction\footnote{We test this hypothesis in \cref{sec:exp_reasoner}.}.
To this end, we train \vlt on track-understanding tasks that connect tracks to scene objects, actions and editing instructions. Such training is necessary because existing VLMs remain limited in reading supplied tracks and relating them to the scene and the instruction. Their ability to output coordinates, for example to localize objects or track points~\cite{llavaov2,molmo2}, does not extend to reading coordinates supplied as input.

\paragraph{Track-understanding tasks.}
\label{sec:vlt_data}

We train \vlt for 40 tasks built around three abilities: (1) \emph{scene grounding}, relating the tracks to objects in the source scene; (2) \emph{motion reading}, recognizing the motion attributes and actions from the tracks; and (3) \emph{edit verification}, judging the agreement between the editing instruction and the source--target track pairs. 
We prepare 1.9M questions, each designed to require the tracks for answering; \cref{fig:overview}b shows one example per ability, and \cref{sec:supp_curriculum} lists all 40 tasks.

Scene-grounding tasks ask which object a tracked point lies on, identified by its point ID, with labels from instance masks, so that each tracked point is tied to an object rather than treated as a bare coordinate.
In the top row of \cref{fig:overview}b, the point with ID 3 lies inside the horse mask and is labeled ``horse''.

Motion-reading tasks ask for motion attributes such as direction, speed and rotation, with labels computed from the tracks, and for a description of the action the tracks express, with labels from captions.
Action questions give only the first frame with the tracks, so later frames cannot reveal the action.
As shown in the middle row of \cref{fig:overview}b, tracks A and B start from the same walking horse but express distinct actions: jumping and rearing up, respectively.

Edit-verification tasks ask whether the target tracks carry out the instruction.
A mismatching input keeps the instruction but takes the target tracks of a different edit of the same source video; the source video, source tracks and point IDs stay fixed, so the two differ only in the target tracks.
Track-prediction tasks train the reverse direction, from the instruction and source context to the motion of individual points.
In the bottom row of \cref{fig:overview}b, target tracks A match ``Change the horse walking to jumping'' but not ``Change the horse walking to rearing up,'' and the reverse holds for tracks B.

\paragraph{Track representation.}
\label{sec:vlt_grammar}
To pose these tasks, we represent source and target tracks as coordinate text so that \vlt can jointly interpret track geometry, scene context and editing intent within a unified multimodal prompt.
Built on Molmo2-8B~\cite{molmo2}, \vlt uses its native coordinate-text format to directly access numerical positions.
For example, \texttt{<tracks coords="0.0 3 450 700; ...">} denotes point ID 3 at $(450,700)$ at timestamp $0.0$ seconds, with coordinates normalized to $[0,1000]$.
Separate \texttt{original} and \texttt{edited} blocks distinguish source and target tracks, while shared point IDs and timestamps establish their correspondence.
Our ablations in \cref{sec:vlt_interface} show that coordinate-text representations outperform rendered trajectories as video and learned track encoders in track understanding.

\paragraph{Training.}
We train a single reasoner on all tasks as a curriculum, a staged mixture whose sampling probabilities change over training.
Early stages emphasize scene-grounding and motion-reading tasks, which tie the tracks to objects and motion, and later stages emphasize edit-verification and track-prediction tasks, which use these readings to relate source motion, target motion and the instruction.
We optimize cross-entropy on answer tokens with LoRA on the attention and feed-forward projections, keeping the backbone weights and the vision encoder frozen.

\subsection{Coupling \vlt to the Generator}
\label{sec:vltgen}

\paragraph{Base generator.}
Following MotionV2V~\cite{motionv2v}, the generator is a pretrained text-to-video diffusion model~\cite{wan21} with a ControlNet branch~\cite{controlnet} for track conditioning.
We render the source and target tracks as two RGB videos of Gaussian blobs, one color per point, drawn only in frames where the point is visible.
These two videos and the source video are VAE-encoded, and their latents are concatenated along the channel dimension as the ControlNet input.

\paragraph{Prompt hidden states.}
On top of this track conditioning, the generator receives prompt hidden states of the frozen \vlt, i.e., hidden states of its input tokens for $(V_s,T_s,T_t,I)$ rather than of its outputs for the track-understanding tasks.
Since the two models differ in depth, width, and token representations, we introduce a coupling that selects the prompt tokens to transfer, learns which reasoner layer supplies each generator block and projects the selected states into attention keys and values.
\Cref{fig:overview}c illustrates these choices within one generator block.

\paragraph{Selecting edit-relevant token spans.}
\label{sec:vltgen_setup}
\label{sec:vltgen_canvas}

We transfer the hidden states of the target-track and instruction token spans (\cref{fig:overview}a, span box), which specify the proposed motion and the editing intent.
Through causal attention, these tokens attend to the preceding source video and source tracks, so their hidden states already incorporate source context and the source spans need not be transferred.
A similarity analysis between reasoner and generator representations supports this choice: hidden states of the two spans align more closely with the generator features of the edited video than those of the full prompt, the two spans with source tracks added, or the instruction span alone (\cref{sec:supp_token_spans}).
The two spans are complementary: each target-track token carries the motion of one point at one time, whereas instruction tokens, which also attend to the target tracks, relate this motion to the requested edit as a whole.
We denote the hidden states of the selected tokens at reasoner layer $l$ by $H^{(l)}$ and transfer them as individual tokens without pooling, so that each track token stays tied to its point and timestamp.

\paragraph{Routing reasoner layers to generator blocks.}
\label{sec:vltgen_route}
Transferring representations between independently pretrained models is challenging because the models differ in depth and in the information encoded at each layer.
Consequently, directly aligning their layer stacks, as in matched-depth architectures~\cite{mot,bagel,cosmos3}, may not provide suitable representations for each generator block.
To address this, we introduce \emph{hard layer routing}, which allows each generator block to select a single reasoner layer while preserving both models' original architectures.
By selecting one layer, the coupling transfers its representations directly to the projection module without averaging features across layers (\cref{sec:exp_vltgen_coupling}).
Each block $b$ learns a vector of selection logits $\alpha^{(b)}$ shared across all edits, rather than an input-dependent router~\cite{moroute}, so the learned mapping is fixed at inference and directly readable.

During training, block $b$ selects from a set of candidate layers $\mathcal L$ spanning the reasoner's depth and a \emph{skip option} $\varnothing$ using $\alpha^{(b)}$ (\cref{fig:overview}c, left):
\begin{equation}
l_b^\ast
=
\arg\max_{l\in\mathcal L\cup\{\varnothing\}}
\left(\alpha_l^{(b)}+g_l^{(b)}\right),
\end{equation}
where $g_l^{(b)}\sim\mathrm{Gumbel}(0,1)$ is sampled independently for each block and candidate.
Block $b$ receives $h^{(b)}=H^{(l_b^\ast)}$, with $H^{(\varnothing)}=0$.
As the argmax is not differentiable, we employ the straight-through Gumbel-Softmax estimator~\cite{gumbelsoftmax,concretedist}.
At inference, each block selects the candidate layer or the skip option with the largest learned logit.

\paragraph{Type-specific projection for joint attention.}
\label{sec:vltgen_project}

To integrate the selected reasoner representations into video generation, each generator block projects them into additional keys and values for \emph{joint attention}~\cite{cosmos3} with video tokens.
However, the two token types differ in kind and in count: track tokens encode coordinates and far outnumber instruction tokens, which encode the edit in language.
A shared projection may therefore favor preserving track-related information at the expense of instruction-specific semantics.
To investigate this, we assess how well edit-dependent feature variations are preserved before and after projection for each token type.
Our preliminary study reveals that shared projections retain less edit-dependent variation for instruction tokens than for track tokens (\cref{sec:exp_vltgen_coupling}).
Motivated by this observation, we propose \emph{type-specific value projections} that independently adapt track and instruction representations while sharing the key projection, which governs attention allocation rather than the content transferred to video tokens.

For reasoner token $p$ of type $\tau(p)\in\{\mathrm{trk},\mathrm{ins}\}$ (target-track or instruction), each block first normalizes its hidden state with $u_p=\mathrm{RMSNorm}(h_p)$.
The normalized states are then projected with a key map shared by both types and a value map chosen by type (\cref{fig:overview}c, middle):
\begin{equation}
k_p = W_k\,u_p, \quad v_p = W_{v,\tau(p)}\,u_p.
\end{equation}
Block indices and bias terms are omitted for brevity.

For rotary positional encoding, each target-track token takes the position on the generator's latent video grid given by the point and time in its coordinate statement, and instruction tokens take positions outside this grid that preserve their order.
The video queries $Q_v$ then attend jointly over the video keys and values $K_v,V_v$ and the reasoner keys and values $K_c,V_c$ (\cref{fig:overview}c, right):
\begin{equation}
\mathrm{Attn}
=
\mathrm{softmax}\!\left(
\frac{Q_v[K_v;K_c]^\top}{\sqrt{d}}
\right)[V_v;V_c],
\label{eq:inject}
\end{equation}
where $[\cdot\,;\cdot]$ denotes concatenation along the token dimension and $d$ is the attention head dimension.

\paragraph{Training and inference.}
We jointly optimize the coupling, generator LoRA parameters, and the ControlNet branch with a flow-matching objective~\cite{flowmatching}, keeping \vlt and the generator base weights frozen.
At inference, we compute the selected reasoner states once per edit and reuse them across denoising steps; the inference cost is reported in \cref{sec:supp_cost}.
Data and implementation details of \vlt and the coupling are provided in the supplementary material.

\section{Experiments}
\label{sec:exp}
\begin{table*}[t]
\centering\small
\setlength{\tabcolsep}{2pt}
\caption{\textbf{Quantitative comparison with track-conditioned video editing methods across four scenarios.}
Best reported results are in \textbf{bold}, and second-best results are \underline{underlined}.
$^{\dagger}$Our reproduction on Wan2.2-A14B.}
\label{tab:main}
\resizebox{\textwidth}{!}{\begin{tabular}{@{}lcccccccccccccccccccc@{}}
\toprule
 & \multicolumn{5}{c}{\begin{tabular}[b]{@{}c@{}}Joint object--camera\\motion editing\end{tabular}} & \multicolumn{5}{c}{\begin{tabular}[b]{@{}c@{}}Only camera motion editing\\in dynamic scene\end{tabular}} & \multicolumn{5}{c}{\begin{tabular}[b]{@{}c@{}}Only camera motion editing\\in static scene\end{tabular}}  & \multicolumn{5}{c}{\begin{tabular}[b]{@{}c@{}}Only object\\motion editing\end{tabular}} \\
\cmidrule(lr){2-6}\cmidrule(lr){7-11}\cmidrule(lr){12-16}\cmidrule(lr){17-21}
Method & EPE$_{\mathrm{c}}\downarrow$ & EPE$_{\mathrm{un}}\downarrow$ & PSNR$\uparrow$ & SSIM$\uparrow$ & LPIPS$\downarrow$ & EPE$_{\mathrm{c}}\downarrow$ & EPE$_{\mathrm{un}}\downarrow$ & PSNR$\uparrow$ & SSIM$\uparrow$ & LPIPS$\downarrow$ & EPE$_{\mathrm{c}}\downarrow$ & EPE$_{\mathrm{un}}\downarrow$ & PSNR$\uparrow$ & SSIM$\uparrow$ & LPIPS$\downarrow$  & EPE$_{\mathrm{c}}\downarrow$ & EPE$_{\mathrm{un}}\downarrow$ & PSNR$\uparrow$ & SSIM$\uparrow$ & LPIPS$\downarrow$ \\
\midrule
TrajAttn~\cite{trajattn} & 43.38 & 42.91 & 12.96 & 0.538 & 0.542 & 36.68 & 35.86 & 13.85 & \underline{0.528} & 0.611 & 39.08 & 40.35 & 14.59 & 0.410 & 0.440  & 23.09 & 23.29 & 15.57 & 0.588 & 0.311 \\
ReVideo~\cite{revideo} & 37.29 & 37.17 & 13.21 & 0.557 & 0.549 & 23.98 & 17.16 & 13.49 & 0.508 & 0.625 & 31.27 & 32.16 & 14.07 & 0.405 & 0.506  & 15.28 & 11.08 & 18.70 & \underline{0.817} & 0.171 \\
MotionV2V$^{\dagger}$~\cite{motionv2v} & 11.75 & 16.32 & \underline{15.26} & \underline{0.575} & \underline{0.393} & 13.37 & 10.91 & 13.41 & 0.473 & 0.546 & 8.69 & 12.05 & \underline{16.70} & \underline{0.498} & \underline{0.271}  & 5.18 & 5.81 & 19.02 & 0.786 & 0.167 \\
Go-with-the-Track~\cite{gowiththetrack} & \underline{10.00} & \underline{13.79} & 13.78 & 0.530 & 0.470 & \underline{12.80} & \underline{10.53} & \underline{14.82} & 0.522 & \underline{0.497} & \underline{7.86} & \underline{11.13} & 15.57 & 0.438 & 0.327  & \textbf{2.73} & \textbf{3.47} & \underline{21.05} & 0.808 & \underline{0.120} \\
\midrule
\textcolor{thinkerBellGreen}{Think}erBell (Ours) & \textbf{6.45} & \textbf{8.71} & \textbf{16.47} & \textbf{0.619} & \textbf{0.327} & \textbf{10.26} & \textbf{8.33} & \textbf{16.29} & \textbf{0.569} & \textbf{0.400} & \textbf{5.53} & \textbf{6.93} & \textbf{21.10} & \textbf{0.639} & \textbf{0.150}  & \underline{3.69} & \underline{4.75} & \textbf{21.62} & \textbf{0.840} & \textbf{0.113} \\
\bottomrule
\end{tabular}}
\end{table*}
\begin{figure*}[t]
\centering
\includegraphics[width=\textwidth]{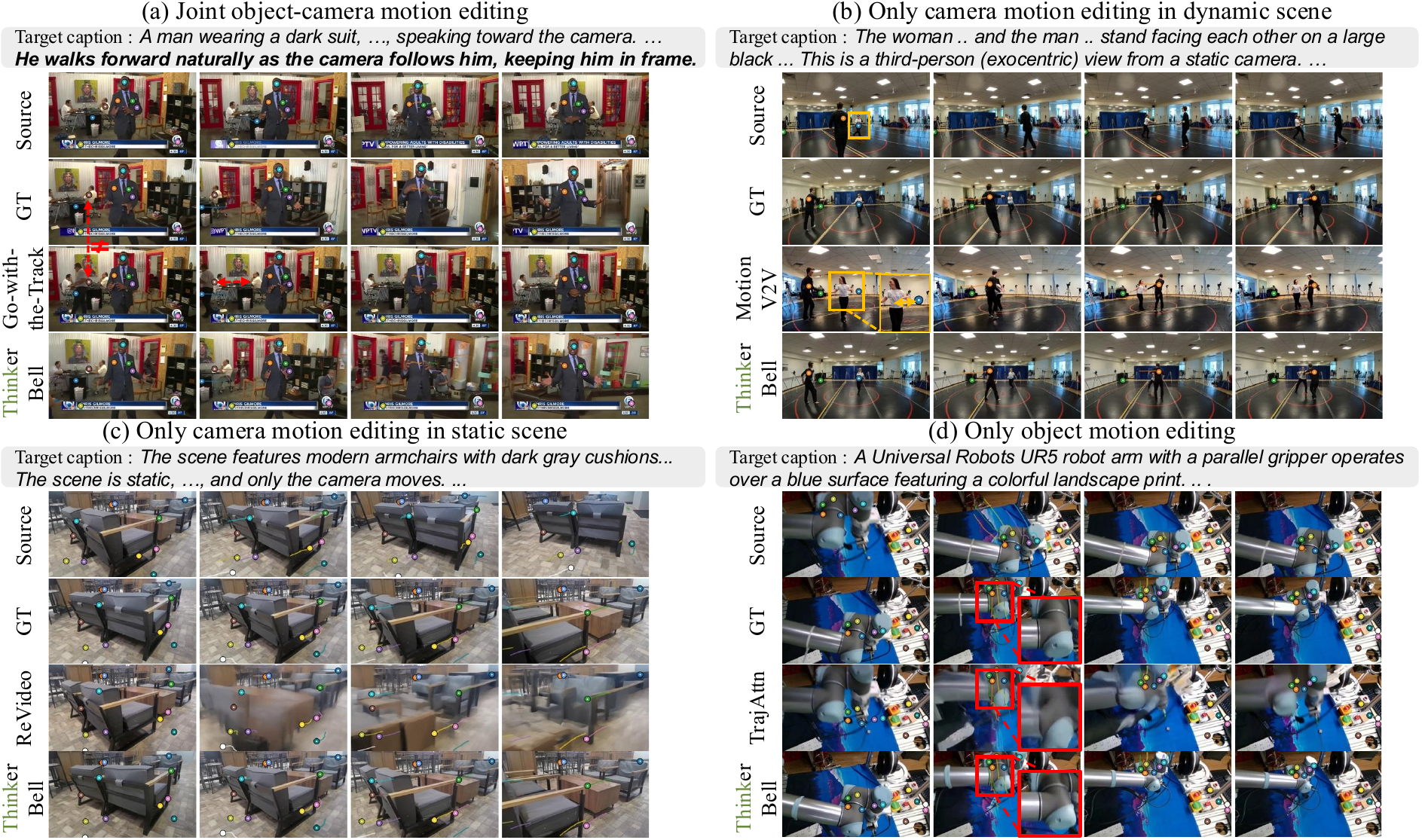}
\caption{\textbf{Qualitative comparison with competing methods across four motion-editing settings.}
Each row shows four frames sampled uniformly at the same time steps.
``Source'' overlays the source tracks on the source video; ``GT'' (the ground-truth target video, not provided to any method) and the generated videos overlay the conditioning target tracks, marking where each point should be.
Points of the same color belong to the same track.
Each tail traces the track over the frames not shown since the previous frame, and no dot is drawn where the conditioning track is marked invisible.
Target descriptions are shown only in part, and editing instructions are omitted.}
\label{fig:qual}
\end{figure*}

\subsection{Experimental Setup}
\label{sec:exp_setting}

\paragraph{Implementation details.}
We evaluate the full \vltgen: \vlt, trained as in \cref{sec:vlt}, is coupled to a Wan2.2-A14B~\cite{wan21} generator for the main comparison.
We train on about 100K source--target video pairs built from OpenVid~\cite{openvid}, with real videos as targets; following MotionV2V~\cite{motionv2v}, each source is generated by Wan2.2-TI2V-5B~\cite{wan21} from an intermediate target frame, which keeps the scene content while changing the motion.
BootsTAPIR~\cite{bootstapir} tracks 64 points per video, with source tracks initialized at the target-track positions in the shared frame to establish correspondence.
Qwen3-VL-30B-A3B~\cite{qwen3vl} writes the editing instructions as ``Change \{source\} to \{target\}'' from both videos, and the target descriptions from the target video.

\paragraph{Competitors.}
We compare \vltgen against state-of-the-art track-conditioned V2V baselines, using the official checkpoints of TrajAttn~\cite{trajattn}, ReVideo~\cite{revideo}, and Go-with-the-Track~\cite{gowiththetrack}.
Since MotionV2V~\cite{motionv2v} is not publicly available, we reimplement it on Wan2.2-A14B and train it on the same data as \vltgen.

\paragraph{Benchmark.}
Existing video motion editing benchmarks are either not publicly available~\cite{motionv2v} or focused on dense 3D trajectories~\cite{editbytrack}, limiting their reproducibility and motion diversity.
We therefore construct a benchmark for sparse 2D track-conditioned video editing\ifarxivversion\else\footnote{The benchmark will be released in the camera-ready version.}\fi{} with four categories of 100 real source--target video pairs each.
To ensure scene consistency, we construct each category from publicly available datasets, pairing videos from the same capture: non-overlapping clips from Koala-36M~\cite{koala} for joint object--camera editing, synchronized views from Harmony4D~\cite{harmony4d} and distinct camera paths from DL3DV-10K~\cite{dl3dv} for camera-only editing in dynamic and static scenes, respectively, and fixed-camera clips from RH20T~\cite{rh20t} for object-only editing.
Since these datasets lack corresponding trajectories, we establish point correspondences by tracking across clips or matching across views using calibrated geometry, retaining 64 tracks per pair that satisfy consistency checks.
For joint object--camera editing, we additionally obtain moving-object masks using RoMo~\cite{romo} and retain only pairs exhibiting substantial trajectory changes in both foreground objects and the background.
Instructions and target descriptions are annotated for each pair as in training.
Beyond this annotation, the target video serves only as the ground truth; no method receives it as input (\cref{sec:supp_benchmark_construction}).

\paragraph{Metrics.}
For each pair, we sample the number of conditioning tracks uniformly as $K\sim\mathcal{U}\{3,\ldots,20\}$ and use the same selected tracks across methods.
For track controllability, we re-track the generated video with CoTracker3~\cite{cotracker3} and report the end-point error (EPE)~\cite{editbytrack}, the mean L2 distance to the ground-truth tracks, with a penalty for visibility mismatches.
EPE$_{\mathrm{c}}$ is measured at the $K$ conditioning points and EPE$_{\mathrm{un}}$ at the remaining $64-K$ ground-truth tracks, which no method receives as input.
With sparse conditions, most of the moving object or region is left unspecified, so EPE$_{\mathrm{un}}$ measures whether a method infers the motion of the whole object or region from the few given tracks, rather than moving only the conditioned points.
For target-video fidelity, we report PSNR, SSIM~\cite{ssim}, and LPIPS~\cite{lpips} against the ground-truth target video.
Implementation and evaluation details are given in \cref{sec:supp_bridge,sec:supp_benchmarks}.

\subsection{Comparison with SOTA Methods}
\label{sec:exp_qual}

\paragraph{Quantitative comparison.}
As shown in \cref{tab:main}, \vltgen achieves the best result in 18 of 20 metrics: all metrics in the three settings with camera motion, and all fidelity metrics in object-only editing, where it is second in the two EPE scores.
Notably, the improvement extends beyond the conditioning tracks: against Go-with-the-Track, the strongest competitor in EPE, \vltgen reduces EPE$_{\mathrm{un}}$, measured at points that no method receives, by 36.8\% in joint object--camera editing and by 20.9\% and 37.7\% in camera-only editing of dynamic and static scenes.
These gains show that \vltgen moves the whole object or region implied by the sparse tracks, which is what scene-aware motion understanding is meant to provide.
\vltgen also improves PSNR by 1.21--4.40\,dB over the strongest competing result in these three settings.

\paragraph{Qualitative comparison.}

\Cref{fig:qual} presents qualitative comparisons, highlighting two major limitations of existing methods: inaccurate motion control (a, b) and poor source-content preservation (c, d).
In \cref{fig:qual}a (red arrow), Go-with-the-Track receives tracks on a background person, displaced to request a camera motion, yet it moves the person while the camera stays still, even though the target description also describes the camera motion.
In \cref{fig:qual}b (yellow arrow), MotionV2V fails to generate the person at the target position when the tracks are placed on a small person in the source video.
In \cref{fig:qual}c and d, ReVideo and TrajAttn produce blurry artifacts when the motion of a large region changes, losing the source identity.
In contrast, \vltgen infers the intended motion even from sparse tracks and preserves the source scene across these challenging edits.
Further comparisons are in \cref{sec:supp_recon,sec:supp_density,sec:supp_i2v}.

\subsection{Analysis}
\label{sec:exp_analysis}
\label{sec:exp_vlt}
\label{sec:exp_vltgen}

We first verify that the track-understanding tasks teach \vlt to understand tracks in context, then test whether this understanding improves generation and whether the improvement holds across generators with different track-conditioning mechanisms.
The generation ablations (\cref{tab:vltgen_reasoner,tab:vltgen_hosts}) use Wan2.1-1.3B~\cite{wan21} as the generator, since training each variant on Wan2.2-A14B is costly; \vlt, the coupling and the evaluation pairs are those of \cref{tab:main}.

\paragraph{Effect of the track-understanding tasks on \vlt.}
\label{sec:exp_curriculum}
\begin{table}[t]
\centering\footnotesize
\setlength{\tabcolsep}{2pt}
\caption{\textbf{Effect of the track-understanding tasks on \vlt.} Chance is the expected score of independent, uniformly random answers under each metric's scoring rule.}
\label{tab:vlt_curriculum}
\resizebox{\columnwidth}{!}{\begin{tabular}{@{}llcccc@{}}
\toprule
 & Model & Pair$\uparrow$ & Read$\uparrow$ & Judge$\uparrow$ & Track gain$\uparrow$ \\
\midrule
 & \color{gray}\textit{Chance} & \color{gray}0.063 & \color{gray}0.113 & \color{gray}0.235 & \color{gray}0.000 \\
\midrule
(a) & VLM~\cite{molmo2} & 0.000 & 0.235 & 0.272 & 0.025 \\
(b) & (a) + editing-only finetuning & 0.010 & 0.253 & 0.265 & 0.045 \\
(c) & (a) + track-understanding tasks (\vlt) & \textbf{0.528} & \textbf{0.831} & \textbf{0.612} & \textbf{0.320} \\
\bottomrule
\end{tabular}}
\end{table}

\Cref{tab:vlt_curriculum} compares \vlt (c) with the pretrained VLM (a) and with a model finetuned only to predict the target tracks from the source video, source tracks and editing instruction (b), which we call editing-only finetuning.
The two finetuned models (b) and (c) see the same inputs and share all training settings; they differ only in their training tasks.

We evaluate the three reasoners independently of the generator, on fixed held-out questions from 15 task types and summarize them in four metrics: \textit{Pair} (one task), \textit{Read} and \textit{Judge} (seven tasks each, averaged, scored with CircularEval~\cite{mmbench}), and \textit{Track gain} (\cref{sec:supp_vlt_metrics}).
\textit{Pair} tests whether the model binds an instruction to a motion change.
Given a source video, its source tracks, and two instructions, describing changes in motion and action, with their target tracks, the model is asked for each of the four instruction--track combinations whether the target tracks carry out the instruction; a pair counts as correct only when both matching combinations are accepted and both mismatching ones are rejected.
\textit{Read} tests whether the model can read motion from tracks in the context of the scene.
In one of its tasks, for example, given only the first frame of a video and its full tracks, the model is asked which object a track belongs to and how it then moves, such as in which direction or with what change in speed.
\textit{Judge} tests whether the model can check a description of motion, stated as a direction or as an action, against the tracks and the video.
In one of its tasks, for example, given the video with its original tracks, an edited set of the same points and an editing instruction, the model is asked whether the edited tracks carry out the instruction.
\textit{Track gain} repeats the Read and Judge questions with the tracks removed from the input and reports the accuracy gained by providing them, averaged over tasks.

\vlt (c) achieves the highest scores on all four metrics, whereas editing-only finetuning (b) barely improves on the pretrained VLM (a).
Although (b) is given the editing instructions and supervised to output the matching target tracks, its Pair score stays near zero: exposure to track--instruction data does not teach the model to relate a motion change to the instruction.
The curriculum instead decomposes this ability into tasks, from reading track geometry and grounding tracks in objects to verifying edits, and (c) acquires all of them, with a Track gain seven times that of (b).
Understanding tracks in context therefore depends on how the training tasks are constructed, not on the data alone.
Further analyses of \vlt are in \cref{sec:vlt_interface,sec:vlt_cert,sec:supp_general}.

\paragraph{Effect of coupling and understanding on generation.}
\label{sec:exp_reasoner}
\begin{table}[t]
\centering\small
\setlength{\tabcolsep}{3pt}\caption{\textbf{Effect of coupling and motion understanding on video generation.}}
\label{tab:vltgen_reasoner}
\resizebox{\columnwidth}{!}{\begin{tabular}{@{}lccccc@{}}
\toprule
Method & EPE$_{\text{c}}\downarrow$ & EPE$_{\text{un}}\downarrow$ & PSNR$\uparrow$ & SSIM$\uparrow$ & LPIPS$\downarrow$ \\
\midrule
\onlygen & 21.02 & 24.74 & 12.65 & 0.499 & 0.532 \\
VLM+Generator& 18.84 & 23.47 & 12.94 & 0.514 & 0.520 \\
VLT+Generator (\vltgen) & \textbf{15.03} & \textbf{19.71} & \textbf{13.07} & \textbf{0.516} & \textbf{0.513} \\
\bottomrule
\end{tabular}}
\end{table}

\Cref{tab:vlt_curriculum} establishes that \vlt understands track motion and editing intent better than the original VLM.
In \cref{tab:vltgen_reasoner}, we test whether this understanding improves generation by comparing \vltgen, i.e., \vlt+Generator, with VLM+Generator, which couples the original VLM to the generator with the same coupling, and with the generator-only baseline; all three share the generator backbone and training data.
Coupling the original VLM already reduces conditioned-point error and improves all three reconstruction metrics over the generator-only baseline, showing that our coupling transfers reasoner features the generator can use.
Replacing the VLM with \vlt yields a larger gain in trajectory control: relative to the generator-only baseline, coupling the original VLM reduces conditioned- and unconditioned-point error by 10\% and 5\%, whereas \vlt reduces them by 28\% and 20\%, 1.7 and 3 times as much.
The two models differ only in the track-understanding training of the reasoner, so this gain is attributable to what \vlt learned from it.
The gain is as large at points that receive no track as at conditioned points, so it extends beyond the motion the user specifies.
Track understanding, rather than the presence of a reasoner, is therefore the main source of \vltgen's improvement in trajectory control.

\paragraph{Generality across generators.}
\label{sec:exp_transfer}
\begin{table}[t]
\centering
\ifarxivversion
\small\renewcommand{\arraystretch}{1.1}
\setlength{\tabcolsep}{3pt}
\def\hoststablebox#1{\resizebox{0.95\columnwidth}{!}{#1}}
\else
\footnotesize\renewcommand{\arraystretch}{0.9}
\setlength{\tabcolsep}{2.4pt}
\def\hoststablebox#1{\scalebox{0.9}{#1}}
\fi
\caption{\textbf{Generality across track-conditioning mechanisms.}}
\label{tab:vltgen_hosts}
\hoststablebox{\begin{tabular}{@{}lccccc@{}}
\toprule
Conditioning & EPE$_{\text{c}}\downarrow$ & EPE$_{\text{un}}\downarrow$ & PSNR$\uparrow$ & SSIM$\uparrow$ & LPIPS$\downarrow$ \\
\midrule
Control branch & 21.02 & 24.74 & 12.65 & 0.499 & 0.532 \\
\quad + \vlt & \textbf{15.03} & \textbf{19.71} & \textbf{13.07} & \textbf{0.516} & \textbf{0.513} \\
\midrule
Latent splatting & 37.76 & 38.93 & 12.40 & 0.497 & 0.552 \\
\quad + \vlt & \textbf{26.99} & \textbf{29.43} & \textbf{13.47} & \textbf{0.529} & \textbf{0.484} \\
\bottomrule
\end{tabular}}
\end{table}

To validate the generality of \vlt and our coupling, we attach them to generators with different track-conditioning mechanisms.
Alongside the control-branch generator in \cref{sec:method}, we evaluate a latent-splatting generator inspired by Edit-by-Track~\cite{editbytrack}, which places source features at target-track positions; both use the same \vlt model and coupling design (\cref{sec:supp_bridge}).
Against its generator-only counterpart (\cref{tab:vltgen_hosts}), \vltgen improves all five metrics with both mechanisms, reducing unconditioned-point error by 20\% with the control branch and by 24\% with latent splatting, so its benefit extends beyond a single track-conditioning mechanism.
Ablations of our coupling design, including layer routing, value projection and token selection, are provided in \cref{sec:exp_vltgen_coupling}.

\FloatBarrier
\section{Conclusion}
\label{sec:conclusion}
Sparse tracks tell a generator where a few points should move, but not what the edit means for the rest of the scene.
We showed that this meaning can be learned by a separate reasoner and handed to a structurally different generator through its internal representations, so that the generator follows sparse tracks more precisely and better preserves the source content; the gain grows with the reasoner's understanding and with how well the coupling transfers it, and holds for both track-conditioning mechanisms.
Understanding before editing thereby establishes a new state of the art in track-conditioned video motion editing.
Our benchmark, with four motion-type categories of diverse real video pairs, reveals where a method succeeds or fails. 
Further discussion and limitations are in \cref{sec:supp_discussion,sec:supp_limitations}.

\section*{LLM Usage Statement}
The manuscript was drafted by the authors and subsequently polished with assistance from OpenAI Codex and Anthropic Claude Code.
These tools also assisted with implementing and debugging experimental code, but were not used to develop research ideas or design experiments.
OpenAI's image-generation tool was used to create and refine the illustrative emoji icons in the title and \cref{fig:teaser}.

\section*{Acknowledgments}
This work was supported by the ``Advanced GPU Utilization Support Program'' funded by the Government of the Republic of Korea (Ministry of Science and ICT) and by the National Research Foundation of Korea (RS-2026-25497603: Fundamentals of Online 4D Reconstruction with Semantic Viewpoint Control and Generative Priors for Proactive Content Consumption).
{
    \small
    \bibliographystyle{ieeenat_fullname}
    \bibliography{main}
}
\clearpage
\begin{bibunit}[ieeenat_fullname]
\clearpage
\appendix
\setcounter{page}{1}
\renewcommand{\thepage}{S\arabic{page}}
\addtocontents{toc}{\protect\setcounter{tocdepth}{2}}
\makeatletter
\newpage
\twocolumn[
  \centering
  \Large
  \textbf{\thetitle}\\
  \vspace{0.5em}Supplementary Material \\
  \vspace{1.0em}
  \ifarxivversion\else
  \begin{minipage}{\textwidth}
    \normalsize\centering
    \textbf{Project Page and Video Results:} \textcolor{red}{\textbf{Please double-click \texttt{ThinkerBell.html}}} to view our project page in a web browser.
    The page presents video comparisons and applications, including edits with user hand-drawn track inputs.
  \end{minipage}\par
  \vspace{1.0em}
  \fi
  \begin{minipage}{\textwidth}
    \normalsize\raggedright
    {\large\bfseries Contents}\par\medskip
    \@starttoc{toc}
  \end{minipage}
  \vspace{1.5em}
]
\makeatother

\clearpage

\section{VLT: Data, Tasks, and Implementation}
\label{sec:supp_vlt}

This section details the input prompt template, track-understanding training tasks, and training setup for \vlt.

\subsection{Track Format and Generation-Time Prompt}
\label{sec:supp_binder}
We represent tracks as coordinate text normalized to $[0,1000]$, with $x$ increasing from left to right and $y$ from top to bottom.
Each time group begins with a timestamp in seconds, followed by triples of point ID, $x$, and $y$; semicolons separate time groups in chronological order.
For example, \texttt{0.0 3 240 630; 0.5 3 280 630} describes point ID 3 moving from $(240,630)$ to $(280,630)$ between 0.0 and 0.5 seconds.
The \texttt{original} and \texttt{edited} blocks contain source and target tracks, respectively, with shared point IDs and timestamps.
The serializer rounds and clips coordinates to the normalized range and does not include a separate visibility field.

At generation time, the reasoner prompt contains the source video, a description of the track format, the source and target track blocks, and the editing instruction followed by a verification request.
We reproduce the format description below:
\begin{quote}\small\raggedright
Coordinates are on a 0-1000 normalized scale (x: 0=left edge, 1000=right edge; y: 0=top edge, 1000=bottom edge of the frame).
Inside \texttt{coords="..."}, each time group is \texttt{"<time-in-seconds> <point-id> <x> <y> <point-id> <x> <y> ..."}, groups are separated by \texttt{";"} in chronological order.
The video, when one is given, always shows the ORIGINAL motion; an edited motion is never shown as video and is given only as coordinates.
A \texttt{<tracks>} block may be marked \texttt{label="original"} or \texttt{label="edited"}; when both appear they describe the SAME physical points, so the same point id means the same point, both blocks use the same time grid, and they can be compared step by step.
\end{quote}
This is followed by \texttt{<tracks label="original" coords="...">} and \texttt{<tracks label="edited" coords="...">}, then:
\begin{quote}\small
Instruction: `\{instruction\}'.\\
The second track set is the intended result of applying this instruction to the first.
Verify that it does, and describe what changes---which objects, which direction, how much, and how each object's physics changes: its speed profile (accelerating, decelerating, or steady), any rotation or scale change, and whether its shape stays rigid or deforms while moving.
\end{quote}
We compute the prompt hidden states in one forward pass and transfer the target-track and instruction spans to the generator without decoding an answer.

\subsection{Task Groups and Data Construction}
\label{sec:supp_curriculum}
\cref{tab:supp_vlt_task_groups} lists the inputs, answers, supervision sources, and sampling weights of all 40 tasks.
The curriculum develops the three abilities introduced in \cref{sec:vlt}: scene grounding, motion reading, and edit verification.
Track prediction trains the reverse relation, from an editing instruction and source context to target motion.
Motion-reading tasks are divided into attributes and actions according to their supervision source.
Two supporting groups aim at enabling coordinate-text reading through synthetic tracks and retain the pretrained pointing ability of Molmo2.
The resulting 1.9M questions are built from 30,203 real OpenVid~\cite{openvid} clips and 48K synthetic track sets.

\begin{table*}[!t]
\centering
\footnotesize
\setlength{\tabcolsep}{3pt}
\renewcommand{\arraystretch}{1.0}
\caption{\textbf{All 40 track-understanding tasks.}
Inputs: $V_s$, source video (eight frames); $F_s$ and $F_t$, first frames of the source and target videos; $T_s$ and $T_t$, source and target tracks; $T$, tracks without an \texttt{original} or \texttt{edited} label (synthetic in the first group); $P_0$, frame-0 positions of the points only; $I$, editing instruction; $D$, motion descriptions from captions (the caption of the source clip, or the captions with which its edits were generated); $p$, point ID; $o$, object name.
Share: sampling weight of the task in the warmup (0--2K), main (2K--14K) and late (14K--26K, including the 22K--26K continuation) phases, as percentages of the phase total.
\#Q: number of questions.}
\label{tab:supp_vlt_task_groups}
\begin{tabular}{@{}p{0.15\textwidth}p{0.28\textwidth}p{0.125\textwidth}p{0.265\textwidth}r@{\hspace{6pt}}c@{}}
\toprule
Task & Ability and knowledge & Input & Answer (label source) & \#Q & Share (\%) \\
\midrule
\multicolumn{6}{@{}l}{\textit{Synthetic track reading} --- reading the coordinate text; labels exact by construction} \\
Still points & Which points do not move & $T$ & Point IDs & 8K & 5.7 / 0.5 / 0.5 \\
Direction change & Two successive motion directions & $T$ & ``left then down'' & 8K & 5.7 / 0.5 / 0.5 \\
Pairwise distance & Two points approach or separate & $T$ & closer / farther & 8K & 3.8 / 0.5 / 0.5 \\
Group pattern & Points converge, spread or move together & $T$ & 3-way word & 8K & 3.8 / 0.9 / 0.9 \\
Displacement & Which point moves farther & $T$ & Point ID & 8K & 4.7 / 0.9 / 0.9 \\
Camera vs.\ object & All points drift together or only some move & $T$ & camera / object & 8K & 3.8 / 0.9 / 0.9 \\
\midrule
\multicolumn{6}{@{}l}{\textit{Pointing retention} --- keeping the pretrained pointing of Molmo2; no tracks} \\
Object pointing & Locating a named object & $F_s$, $o$ & Molmo2 point (mask centroid) & 59K & 7.5 / 2.7 / 2.7 \\
Object listing & Naming the main objects & $F_s$ & Up to five object names (masks) & 30K & 4.7 / 1.8 / 1.8 \\
\midrule
\multicolumn{6}{@{}l}{\textit{Scene grounding} --- tying point IDs to scene objects} \\
Points on object & Object $\to$ the points on it & $F_s$, $P_0$, $o$ & Point IDs (instance masks) & 57K & 3.8 / 1.8 / 1.8 \\
Object of point & Point $\to$ the object it lies on & $F_s$, $P_0$, $p$ & Object name (instance masks) & 57K & 2.8 / 1.4 / 1.4 \\
Instance identity & Two instances of one category & $F_s$, $P_0$ & Point IDs on the left one (masks) & 1K & 1.4 / 0.9 / 0.9 \\
Part motion & Point moves with its object or on its own & $V_s$, $T$, $p$, $o$ & with / independently (vs.\ object centroid) & 19K & 2.8 / 2.7 / 2.7 \\
Occlusion & Which point becomes hidden & $V_s$, $P_0$ & Point ID (tracker visibility) & 27K & 1.4 / 0.9 / 0.9 \\
\midrule
\multicolumn{6}{@{}l}{\textit{Motion reading: attributes} --- motion properties that direction alone cannot answer; labels computed from the tracks} \\
Object direction & Direction of an object, including depth & $F_s$, $T_s$, $o$ & One of six directions or stationary & 59K & 5.7 / 2.7 / 2.7 \\
Object motion & Direction and amount per object & $F_s$, $T_s$ & One templated sentence per object & 24K & 8.5 / 7.3 / 7.3 \\
Target object motion & The same for the target motion & $F_t$, $T_t$ & One templated sentence per object & 24K & 2.8 / 2.7 / 2.7 \\
Path shape & Straight, curved or back-and-forth path & $F_s$, $T_s$, $p$ & 3-way choice & 73K & 2.8 / 1.8 / 1.8 \\
Speed profile & Accelerating, steady or decelerating & $F_s$, $T_s$, $p$ & 3-way choice & 74K & 2.8 / 1.8 / 1.8 \\
Path contrast & Two same-direction points, different paths & $F_s$, $T_s$ & 2-way choice & 40K & 1.9 / 1.8 / 1.8 \\
Rotation & Object rotation & $F_s$, $T_s$, $o$ & clockwise / counter-clockwise / none & 15K & 0.9 / 1.4 / 1.4 \\
Scale change & Object approaching or receding & $F_s$, $T_s$, $o$ & larger / smaller / same & 16K & 0.9 / 1.4 / 1.4 \\
Rigidity & Object keeps its shape or deforms & $F_s$, $T_s$, $o$ & rigid / deforms (pairwise distances) & 26K & 0.9 / 1.4 / 1.4 \\
Track--video check & Tracks follow the motion in the video & $V_s$, $T$ & Yes / No (reversed or target tracks) & 116K & 2.8 / 3.7 / 3.7 \\
Motion sentence check & Whether a motion sentence, the tracks and the video agree & $V_s$, $T$, sentence & 3-way choice (corrupted sentence or tracks) & 73K & 0 / 1.8 / 1.8 \\
\midrule
\multicolumn{6}{@{}l}{\textit{Motion reading: actions} --- what the objects do; labels from captions} \\
Action description & Tracks $\to$ one action sentence & $F_s$, $T_s$ or $T_t$ & Free-form sentence (caption clause) & 84K & 2.8 / 4.6 / 4.6 \\
Action selection & Which action $T_t$ expresses & $F_s$, $T_t$, 3 $D$ & Letter (clauses of sibling edits) & 62K & 1.9 / 2.7 / 2.7 \\
Clip motion caption & Tracks $\to$ caption motion sentence & $F_s$, $T_s$ & Free-form sentence (caption) & 25K & 4.7 / 4.6 / 4.6 \\
Source action match & Which description $T_s$ expresses & $F_s$, $T_s$, 3 $D$ & Letter (captions) & 25K & 2.8 / 5.5 / 4.6 \\
Target action match & Which description $T_t$ expresses & $V_s$, $T_s$, $T_t$, 3 $D$ & Letter (captions of the edits) & 76K & 0 / 6.4 / 5.5 \\
Target action check & Whether a description fits $T_t$ & $V_s$, $T_s$, $T_t$, $D$ & Yes / No (captions of the edits) & 152K & 0 / 5.5 / 5.5 \\
\midrule
\multicolumn{6}{@{}l}{\textit{Edit verification} --- relating the source-to-target change to the instruction} \\
Instruction verification & Whether $T_t$ carries out $I$ & $V_s$, $T_s$, $T_t$, $I$ & Yes / No (tracks of another edit) & 103K & 2.8 / 3.7 / 3.7 \\
Instruction selection & Which instruction the edit follows & $V_s$, $T_s$, $T_t$, 3 $I$ & Letter & 32K & 0 / 2.7 / 2.7 \\
Instruction inference & Writing the instruction of an edit & $V_s$, $T_s$, $T_t$ & Instruction naming the objects & 41K & 2.8 / 5.5 / 5.5 \\
\midrule
\multicolumn{6}{@{}l}{\textit{Track prediction} --- instruction $\to$ motion of individual points; answers from the target tracks} \\
Affected direction & Direction of $p$ after the edit & $V_s$, $T_s$, $I$, $p$ & 5-way choice & 32K & 0 / 6.4 / 4.6 \\
Direction and amount & Direction and amount of $p$ & $V_s$, $T_s$, $I$, $p$ & Direction and amount, or not affected & 152K & 0 / 3.7 / 1.8 \\
Endpoint & Final position of $p$ & $V_s$, $T_s$, $I$, $p$ & Coordinate ``$x$ $y$'' & 78K & 0 / 4.6 / 2.7 \\
Single trajectory & Full target trajectory of $p$ & $V_s$, $T_s$, $I$, $p$ & Track block & 78K & 0 / 0 / 2.7 \\
Reasoned trajectory & The same, after naming the affected object & $V_s$, $T_s$, $I$, $p$ & Reasoning, then track block & 78K & 0 / 0 / 2.7 \\
Multi-point trajectory & Full target trajectories of 2--4 points & $V_s$, $T_s$, $I$, $p$'s & Track block & 38K & 0 / 0 / 0.9 \\
Reasoned multi-point & The same, after naming affected objects & $V_s$, $T_s$, $I$, $p$'s & Reasoning, then track block & 38K & 0 / 0 / 0.9 \\
\bottomrule
\end{tabular}
\end{table*}

\paragraph{Source and target tracks.}
For VLT training, each real clip is the source, and its generated variants provide alternative target motions.
These variants share the scene content but follow different motion from an intermediate frame of the real clip.
They are the same video pairs used for generator training (\cref{sec:exp_setting}), with the source and target roles reversed: generator training uses the generated video as the source and the real video as the target.
Tracking corresponding physical points establishes shared point IDs and timestamps across the two track blocks.
Instance masks provide object identities, track statistics provide motion-attribute labels, and the source caption and generation captions provide action descriptions.
Instruction-based tasks use validated editing instructions, and object-referential tasks use masks that pass quality checks.

\paragraph{Action-centered supervision.}
After 2K iterations, action tasks account for approximately 27--29\% of the sampling weight, which ask the reasoner to describe, select, or verify the action expressed by the tracks.
Candidate descriptions refer to the same objects and background, making motion necessary to distinguish them.
Editing instructions also use actions identified by Qwen3-VL to express changes in what the objects do.
Up to 22K iterations, an action phrase replaces the corresponding track-derived motion phrase when the source and target actions differ, making the action change explicit in the instruction.
When the actions are the same, the instruction retains the motion phrase to describe changes in motion direction or magnitude.
However, when an instruction contains only an action phrase, sparse tracks may not provide directly checkable evidence for that action.
The 22K--26K continuation therefore retains the track-derived motion phrase alongside the action phrase, as in ``the woman moving down steadily while touching her hair.''
This provides motion evidence that can be checked against the tracks while retaining the intended action change.
Action phrases occur in 43\% of instruction-verification questions in the initial corpus and 33\% in the revised corpus.

\paragraph{Inputs that require the tracks.}
Tasks that read source-motion attributes or actions receive the first frame and the tracks.
The frame supplies object appearance and location, while the tracks supply motion.
Tasks that compare tracks with observed video motion or verify an edit receive the eight-frame source video.
To mitigate shortcut learning, we construct negative examples and distractors.
Inconsistent tracks either reverse each point's motion about its starting position, preserving the initial point locations, or come from another edit of the same clip.
Distractor descriptions are complete sentences in the same style as the correct answer, and binary labels are balanced per object.
Real-video track blocks contain 6--12 points and 8--12 time steps sampled per clip, exposing the reasoner to different input lengths.

\paragraph{Edit verification and track prediction.}
Instruction-verification questions pair the source video and source tracks with an instruction and candidate target tracks.
Matching examples use the tracks of the instructed edit; mismatching examples use a different edit of the same source, selected from edit pairs marked as usable negatives during data validation.
Track-prediction tasks query both affected points and points whose motion remains unchanged, with endpoint and full-trajectory answers taken from the target tracks.

We split video-based questions by source clip, keeping all questions and alternative edits of a clip in the same split.
The validation set contains designated clips and a further 2\% selected by a deterministic hash of dataset and clip identifiers.
Synthetic questions are used only for \vlt training.
Task weights specify the requested sampling proportions within each phase; each task pool is capped at three repetitions per sampling epoch to limit oversampling of small pools.

\subsection{Training Schedule and Parameters}
\label{sec:vlt_recipe}
We initialize \vlt from Molmo2-8B and optimize cross-entropy on answer tokens.
LoRA is applied to the attention and feed-forward projections with rank 64 and scaling parameter 64; the backbone weights and vision encoder remain frozen.
We use a learning rate of $10^{-4}$ and a prompt budget of 6,144 tokens.
Video-input tasks use up to eight frames, while first-frame tasks use a single frame.

Training uses warmup (0--2K), main (2K--14K), and late (14K--26K) phases.
The warmup emphasizes synthetic track reading and motion reading, which together receive approximately 70\% of the sampling weight, alongside pointing retention and scene grounding.
The main phase introduces affected-direction, direction-and-amount, and endpoint prediction, and increases edit-verification weight from approximately 6\% to 12\% by adding instruction selection and increasing the weights of the existing tasks.
The late phase adds single- and multi-point trajectory prediction, each with a direct variant and a reasoned variant that names the affected objects before predicting the tracks.
All task groups except track prediction are active from the warmup onward.
At 22K iterations, we resume both model and optimizer states with the revised instruction corpus and retain the late-phase weights through 26K.

\section{\vltgen: Training Data and Implementation}
\label{sec:supp_bridge}

This section provides details on \vltgen: its training data construction, \vlt feature transfer to the generator, and training setup for the generator and coupling modules.
We first describe the Wan2.2-A14B model used in the main comparison (\cref{tab:main}), then specify the layer-routing initialization for the Wan2.1-1.3B generation ablations.

\subsubsection*{Training Data Construction}
\paragraph{Pair construction.}
Following MotionV2V~\cite{motionv2v}, each real OpenVid~\cite{openvid} clip serves as the edited video $V_t$, and we generate source videos $V_s$ that share its content but differ in motion.
Each pair is built in six steps.
\begin{enumerate}[leftmargin=*,label=(\arabic*)]
\item \textit{Target video and description.}
$V_t$ is the first 81 frames of the clip at 16\,fps.
Qwen3-VL-30B-A3B~\cite{qwen3vl} captions 32 uniformly sampled frames of $V_t$ in 80--150 words, giving the target description $C$.
\item \textit{Source descriptions.}
From $C$ alone, the same model writes three source descriptions $C_s$.
Each preserves the descriptions of subject appearance, scene content, and camera movement word for word, while replacing the object motion with a different, physically plausible motion.
\item \textit{Target tracks.}
BootsTAPIR~\cite{bootstapir} tracks 64 points, sampled uniformly at random over frame 0 of $V_t$, through all 81 frames, giving $T_t$.
\item \textit{Source video.}
We draw a branch frame $k$ uniformly from $\{1,\ldots,40\}$ and use frame $k$ of $V_t$ as the first frame of $V_s$.
Wan2.2-TI2V-5B~\cite{wan21} generates the 81-frame $V_s$ at 16\,fps from this frame and one $C_s$, with a guidance scale of 5.5 at $1280\times704$.
Each of the three source descriptions yields one $V_s$, so one target video gives three pairs.
$V_t$ keeps all 81 frames; frame 0 of $V_s$ shows the content of frame $k$ of $V_t$.
\item \textit{Source tracks.}
The positions of the 64 target points at frame $k$, rescaled to the resolution of $V_s$, become the queries at frame 0 of $V_s$, and BootsTAPIR tracks them through $V_s$, giving $T_s$.
Thus, the $j$-th source and target tracks follow the same physical point, with correspondence established between source frame 0 and target frame $k$.
A point invisible at target frame $k$ has no source query and is marked invisible throughout $V_s$.
\item \textit{Editing instruction.}
Qwen3-VL-30B-A3B views each video separately and states the action of each named object and the camera motion; a second pass over different frames rejects statements the video does not support.
The instruction $I$ takes the form ``Change \{source motion\} to \{target motion\}'': each moving object keeps the motion phrase measured from its tracks and adds its action with ``while'' when the actions in $V_s$ and $V_t$ differ, and a camera clause appears when the camera motions differ (\cref{sec:supp_curriculum}).
\end{enumerate}
Qwen3-VL then removes pairs in which the moving object of $V_s$ is not the same instance as that of $V_t$.

\paragraph{Training data.}
We train on 107K source--target pairs built from OpenVid~\cite{openvid}, each with an editing instruction.
Camera-only pairs instead describe the change of camera path or viewpoint, and fixed-camera pairs end with ``while the viewpoint remains unchanged.''
Each training sample conditions the generator on 5--64 tracks, 70\% of them on moving points.

\subsubsection*{Training Details}
\paragraph{Reasoner input.}
We freeze the final \vlt checkpoint described in \cref{sec:supp_vlt}.
It receives eight source-video frames, the editing instruction, and at most 12 source--target track pairs in the format of \cref{sec:supp_binder}; the generator receives all conditioning tracks.
We first select up to six points whose source and target tracks differ, then fill the remaining slots with moving points followed by static points, preferring points visible in the first frame.
The coupling transfers hidden states from the target-track and instruction spans, averaging 742 tokens per edit.

\paragraph{Generator and trainable parameters.}
Wan2.2-A14B has two experts of 40 blocks each; the high-noise expert denoises timesteps $t\ge0.875$ and the low-noise expert the rest.
We train the two experts as separate runs.
Each expert has its own ControlNet branch, a copy of all 40 blocks initialized from the base weights, together with rank-64 LoRA on its attention layers and its own coupling.
Per expert, the trainable parameters are 15.1B in the ControlNet branch, 0.21B in LoRA, 2.52B in the coupling projections and 320 routing logits.
The base weights, VAE, text encoder and \vlt remain frozen.
The ControlNet input concatenates the latents of three VAE-encoded videos (48 channels): the source video and the source and target track videos, in which each point is a Gaussian blob ($\sigma=10$\,px) of one color shared across the two track videos.

\paragraph{Optimization.}
We use AdamW with fp32 master weights ($\beta_2=0.95$, $\epsilon=10^{-12}$), gradient clipping at 0.05 and a global batch of 16 clips of 81 frames at $480\times832$.
Learning rates are $10^{-5}$ for LoRA and the ControlNet blocks, $10^{-4}$ for the ControlNet output projections and the coupling projections, $10^{-3}$ for the ControlNet input embedding, and $10^{-2}$ for the routing logits.
After a 2K-iteration warmup, all rates are multiplied by 0.1 at 8K iterations and by 0.1 again at 11,250 (high-noise) and 12,500 (low-noise) iterations.
The high-noise run samples $t$ uniformly in $[0.875,1)$, and the low-noise run samples $t$ from the distribution shifted with a flow shift of 5, restricted to $t<0.875$.
We use the checkpoints at 12,750 (high-noise) and 13,750 (low-noise) iterations.
At inference, we use 50 UniPC steps with a flow shift of 5 and a guidance scale of 3.0 for both experts; each expert uses its own ControlNet branch and coupling.

\paragraph{Seeded layer routing.}
The candidate reasoner layers are $\{7,12,17,22,27,32,36\}$, together with the skip option.
We map the 40 blocks of each expert proportionally onto layers 7--36 and initialize the logit of the candidate nearest to the mapped layer to 2 and all other logits to zero.
This initializes different blocks with preferences for different reasoner depths while allowing every block to select any candidate during training.
The straight-through estimator uses a temperature of 0.5.

\Cref{sec:supp_routing} analyzes the learned routing.

\paragraph{Projection parameterization.}
Each block applies RMS normalization without affine parameters to the selected reasoner states.
The key projection shares its weight across token types and adds a type-specific bias; the value projections use separate weights and biases for target-track and instruction tokens.
The value projections are zero-initialized, so reasoner tokens initially contribute zero values, while their keys participate in attention normalization.
The equations in the main text omit the projection biases.
Selecting the skip option sets the normalized reasoner states to zero; the projection biases still produce keys and values, so this option removes input-dependent features without disabling the coupling.

\paragraph{Joint attention.}
Joint attention places target-track and video tokens on the same latent video grid through rotary positional encoding (\cref{sec:vltgen_project}).
Each target-track token receives the position specified by its point and timestamp, making this correspondence available to video queries alongside token content.

\subsubsection*{Generator Variants}
\paragraph{Wan2.1-1.3B models.}
The generation ablations (\cref{tab:vltgen_reasoner,tab:vltgen_hosts,tab:vltgen_ablation}) use Wan2.1-1.3B, which has 30 generator blocks compared with 40 per expert in Wan2.2-A14B.
We retain the same coupling design and adjust only the layer-routing initialization to accommodate this difference in depth.
For zero-based block index $b\in\{0,\ldots,29\}$, we initialize the logit of the candidate nearest to reasoner layer $7+b$ to 2 and all other logits to zero.

\paragraph{Latent-splatting variant.}
The latent-splatting variant in \cref{tab:vltgen_hosts} incorporates the source video in a manner similar to Edit-by-Track~\cite{editbytrack}.
The source video is encoded by the same VAE as the target video and converted into tokens.
For track conditioning, we use latent splatting, as in Edit-by-Track, ATI, and Wan-Move~\cite{editbytrack,ati,wanmove}.
We query features from the source-video tokens at the source-track positions.
These features are added back to the source-video tokens at their original positions; the same features are also moved to the corresponding target-track positions and added to the noisy target-video tokens.
We then concatenate the augmented source and target sequences along the token dimension and feed them jointly into the DiT.
The source-video tokens pass through the DiT at timestep $t=0$, while the target-video tokens use the current denoising timestep.

\section{Evaluation Protocols}
\label{sec:supp_eval}
\label{sec:supp_ci}
This section defines the \vlt evaluation metrics and describes the video benchmark construction, video evaluation protocols, and competitor settings.

\subsection{VLT Evaluation Metrics}
\label{sec:supp_vlt_metrics}
We evaluate all models on a fixed set of held-out questions.
For \emph{Pair}, we use 309 pairs of edits, each sharing a source video and containing two instructions with their corresponding target tracks.
Crossing the instructions and tracks gives four yes/no questions: two matching and two mismatching combinations.
A pair counts as correct only when all four judgments are correct.
This strict criterion requires the model to distinguish both different tracks paired with the same instruction and different instructions paired with the same tracks.
A model that consistently bases its judgments solely on the instruction or solely on the tracks cannot correctly classify all four combinations and therefore receives no credit for the pair.
Independent random yes/no answers therefore have an expected score of $1/16$, whereas always answering yes or always answering no scores zero.
For \emph{Read} and \emph{Judge}, we use 300 held-out questions per task across seven tasks each, giving 2,100 questions per metric and 4,200 in total.
We average per-task accuracy with equal weight; cyclic option rotations are repeated evaluations of the same questions and do not increase these question counts.
To reduce answer-position bias, we follow CircularEval~\cite{mmbench}, counting a multiple-choice question as correct only when the model answers it correctly under every cyclic ordering of the options.
For \emph{Track gain}, we evaluate the same questions with and without track input, using a single pass with the original option order in both conditions.
We compute the accuracy difference within each task and average over the fourteen Read and Judge tasks.
Thus, Track gain uses single-pass accuracies rather than the CircularEval accuracies reported for Read and Judge.
The chance row gives the expected scores of independent random answers under each metric's scoring rule, with zero expected Track gain.

\subsection{Video Benchmark Construction}
\label{sec:supp_benchmark_construction}
\begin{table*}[t]
\centering
\footnotesize
\setlength{\tabcolsep}{4pt}
\caption{\textbf{Benchmark construction.}
Each pair consists of two real videos from the same capture; the target video is the ground truth for editing the source video.
Pairs: number of evaluated pairs per category.}
\label{tab:supp_benchmark}
\begin{tabular}{@{}p{0.13\textwidth}p{0.13\textwidth}p{0.29\textwidth}p{0.31\textwidth}r@{}}
\toprule
Category & Data & Source and target videos & Track correspondence & Pairs \\
\midrule
Joint object--camera & Koala-36M~\cite{koala} & Two non-overlapping windows of one clip; the target is the first & One CoTracker3 pass over both windows & 100 \\
Object only & RH20T~\cite{rh20t} & Two adjacent windows of one robot episode, from one fixed camera & One CoTracker3 pass over both windows & 100 \\
Camera only, dynamic & Harmony4D~\cite{harmony4d} & One time window from two synchronized cameras & Cross-view matching, then CoTracker3 per view; background from 3D points & 100 \\
Camera only, static & DL3DV-10K~\cite{dl3dv} & Two overlapping segments of one camera path; the source is played in reverse & One CoTracker3 pass over both segments & 100 \\
\bottomrule
\end{tabular}
\end{table*}
Each category pairs two real videos from the same capture, so the target video shows the source content under the target motion (\cref{tab:supp_benchmark}).
Every pair has 64 ground-truth tracks, built so that the $j$-th source and target tracks follow the same physical point.
All videos have 81 frames stored at 16\,fps, and all distances below are measured at $832\times480$.

\paragraph{Joint object--camera editing (Koala-36M).}
Koala-36M splits its videos at scene transitions, and we take both videos of a pair from one clip.
The target is frames 0--80 of the clip, and the source is a later window of 81 frames that starts at frame 97 or later, leaving a gap of at least 16 frames; we take the first such window, in steps of 8 frames, that passes all checks below.
We sample 64 query points at target frame 0, 45 inside the moving-object mask of RoMo~\cite{romo} and 19 outside it, spread over an $8\times8$ grid.
CoTracker3~\cite{cotracker3} tracks them once from frame 0 through both windows, and the tracks are cut into the two windows.
If fewer points pass on one side, the other side fills the 64, with at least 23 inside and 10 outside.
A track is kept only if it is visible in the first frame of both windows and in at least 40\% of the frames of each window, returns within 6\,px when re-queried at source frame 0 and tracked back, does not collapse onto another track, and does not stay fixed on the screen while the background moves, as on logos and other overlays.
To ensure that both motions change, we compare the two windows in two quantities: the camera change, from the background tracks, and the object change, from the moving-object tracks after removing a per-frame affine camera motion fitted to the background.
A pair is kept when both changes exceed 20\,px and the object change exceeds three times the same measure computed on the background tracks, which reflects parallax and fitting error.
Of 635 clips with at least 178 frames, 367 have a nonempty RoMo mask and 105 yield a candidate pair, from which we select 100 for evaluation.
Across the candidate pairs, the median camera change, object change and background measure are 87, 82 and 6.7\,px.

\paragraph{Object-only editing (RH20T).}
We use robot demonstrations from the fixed cameras of RH20T, Franka and UR5 arms, in its test scene, excluding wrist cameras and cameras mounted on their side.
Each pair splits a gap-free block of 162 frames, taken before the person resets the scene, into two adjacent windows of 81 frames, and a random draw per block decides which window is the source.
We keep blocks in which the gripper moves at least 3\,cm within a window and take at most two blocks per episode.
Frames are kept at the recorded rate of 8--14\,Hz and stored at 16\,fps.
CoTracker3 tracks the whole block once, and the tracks are cut into the two windows.
The 64 points consist of up to 26 moving points and at least 38 static points: up to six on the gripper, located by projecting its recorded position, up to ten on objects the gripper moves, segmented with SAM3, and the rest on the arm; static points come from a grid away from the moving regions.
A track is kept if re-tracking from the source first frame agrees within a median of 2\,px and a 90th percentile of 6\,px, and, except on the gripper, if it is visible in at least 60\% of the frames of each window.
A camera is discarded when fewer than three gripper points pass, when a moved object has no passing point, or when the median difference in point displacement between the two windows is below 10\,px.
We select 100 of the 283 resulting pairs.

\paragraph{Camera-only editing in dynamic scenes (Harmony4D).}
Both videos show the same 81-frame window of a Harmony4D sequence from two synchronized cameras, so the people perform the same motion from two viewpoints.
A camera is either a static exocentric camera or the head-worn egocentric camera of a participant, undistorted to a pinhole view with a 90$^{\circ}$ horizontal field of view; pairs are exocentric to exocentric, egocentric to exocentric, or exocentric to egocentric.
We use an egocentric camera only in windows where its calibrated pose lies within 0.3\,m of the wearer's head, and exclude four sequences whose body fits do not match the video.
Background points are static 3D points from COLMAP, projected into both views.
Up to 32 of the 64 points are person points, and the rest are background points; person points are placed on body parts chosen from the SMPL fit, matched across views with UFM under round-trip and epipolar checks, and tracked by CoTracker3 in each view.
Background tracks must agree with their projections, and the two tracks of a person point must agree with the epipolar geometry.
We exclude pairs without a person point, which leaves 341 test pairs, and select 100 of them: 77 exocentric to exocentric, 12 egocentric to exocentric and 11 exocentric to egocentric.

\paragraph{Camera-only editing in static scenes (DL3DV-10K).}
From a DL3DV-10K video resampled at 32\,fps, we take 121 consecutive frames at $1920\times1080$.
The target is the last 81 frames in forward order, and the source is the first 81 frames in reverse order, so frame 0 of the source is frame 40 of the target and the source camera moves along the path in the opposite direction.
The first 41 frames of the target therefore appear, in reverse order, as the first 41 frames of the source.
CoTracker3~\cite{cotracker3} tracks the 121 frames once from their middle frame, and the tracks are cut into the two videos.
Among 512 candidate points on an $8\times8$ grid, we keep 64 with the highest visibility in both videos, preferring one per grid cell, and discard pairs whose mean visibility in either video is below 0.6.
We use one window per test scene, 289 pairs in total, and select 100 of them.

\paragraph{Descriptions and instructions.}
Qwen3-VL-30B-A3B~\cite{qwen3vl} writes each target description from 32 frames of the target video, as in training.
For RH20T, Harmony4D and DL3DV, the prompt also states facts the frames do not reliably reveal: the robot, the fixed camera and the amount of arm motion; the names of the people and the target viewpoint; and the measured camera motion.
RH20T and Harmony4D descriptions end with a sentence stating the viewpoint, such as ``This is a third-person (exocentric) view from a static camera.''
Instructions are assembled by rule as ``Change \{source\} to \{target\}'':
\begin{itemize}[leftmargin=*]
\item \textit{Koala-36M and RH20T.} Each object clause states the motion measured from its tracks, relative to the camera for Koala-36M, and adds the action that Qwen3-VL observes in each video with ``while'' when the two actions differ, as in training (\cref{sec:supp_bridge}). Objects are named from their SAM3 masks. Koala-36M instructions add a camera clause from the background tracks, and RH20T instructions end with ``while the viewpoint remains unchanged.''
\item \textit{Harmony4D.} ``Change the view from \{source\} to \{target\}, while the people's actions remain unchanged.'' The target view is described relative to the source view, by its rotation around the people and, when large, its change in height or distance; an egocentric view names its wearer and the camera motion.
\item \textit{DL3DV-10K.} ``Change \{source camera motion\} to \{target camera motion\}, while the scene remains unchanged.'' Directions and amounts come from a similarity transform fitted to the tracks every eight frames, and the verb, such as panning or moving, from the calibrated camera pose. Pairs with identical source and target phrases are excluded.
\end{itemize}

\paragraph{Pair selection and ground truth.}
Within each category, we select 100 pairs in the order of a hash of their identifiers, without looking at any model output.
For each pair, the number of conditioning points is drawn as $K\sim\mathcal U\{3,\ldots,20\}$, and the $K$ points are fixed across methods.
They are stratified by motion: 70\% inside the moving-object mask for Koala-36M, half on moving points for RH20T, and up to 70\% on people for Harmony4D, split evenly between two people; DL3DV-10K points are drawn uniformly.
Because the two videos of a pair share the scene, its objects and its lighting, the target video shows the source content under the target motion, and fidelity to it reflects how well the source content is preserved.
Regions that only the target video shows, such as those that the camera motion reveals, cannot be inferred from the source; there, fidelity reflects plausible synthesis rather than preservation.

\subsection{Video Evaluation Protocols}
\label{sec:supp_benchmarks}
\paragraph{Conditioning-track jitter.}
Following the inference protocol of MotionV2V~\cite{motionv2v}, we add uniform jitter of up to $\pm2$ pixels at $480\times832$ to the target tracks provided as conditioning to every method, including \vltgen, drawing the noise independently for each point and frame.
Frame 0 and the source tracks are left unperturbed, and all methods receive the same noise.
We evaluate the generated videos against the original, unperturbed ground-truth tracks.

\paragraph{Visibility-penalized trajectory error.}
We evaluate how accurately the motion in the generated video follows the target trajectories.
We compare the ground-truth target tracks with tracks obtained by re-tracking the generated video.
To obtain these tracks, we place query points in the first frame of the generated video at the first-frame coordinates of the ground-truth target tracks, including both conditioning and withheld points, and track them through the video with CoTracker3.
We compute the L2 distances between the re-tracked and ground-truth positions at matching time steps.
We report EPE$_{\mathrm{c}}$ at the $K$ conditioning points and EPE$_{\mathrm{un}}$ at the remaining $64-K$ points, which are withheld from every method.
At the evaluation resolution of $480\times832$, L2 distances are capped at 52 pixels.
Visibility is estimated by CoTracker3 in both the generated and target videos; a mismatch in either direction receives the same 52-pixel penalty.
Points invisible in both videos are excluded, as are visibility mismatches within one target frame of a reference visibility transition.

\subsection{Competitor Settings}
\label{sec:supp_competitors}
All methods receive the same source--target pairs and the same $K$ conditioning points.
No method receives a frame of the target video $V_t$; the V2V methods see only the source video $V_s$.
Outputs are resized to $480\times832$, and methods that generate fewer than 81 frames are extended to 81 frames by repeating, at each target time step, the generated frame nearest in time.
Their input frames and tracks are taken from $V_s$, $T_s$ and $T_t$ at evenly spaced time steps.

\paragraph{MotionV2V.}
We reimplement MotionV2V on Wan2.2-A14B, with a separate ControlNet for each of the high- and low-noise experts.
It receives $V_s$, $T_s$, $T_t$ and $C$, that is, the inputs of \vltgen without the instruction $I$, and generates 81 frames.

\paragraph{Go-with-the-Track.}
We use the official 480p checkpoint on Wan2.2-A14B, which generates 49 frames at $480\times832$ from the description $C$.
The model takes up to four reference frames, the track positions on them, and the tracks over the generated frames.
We sample four frames of $V_s$ at random as references, give $T_s$ at those frames, and give $T_t$ resampled at the 49 generated time steps.
The model accepts 2D tracks, so no depth is provided.
Points that are not visible in any reference frame are passed as tracks without a reference position, which the model supports.
Go-with-the-Track is trained with 256 to 16{,}384 tracks and evaluated with at least 32, whereas our benchmark provides 3 to 20.

\paragraph{TrajAttn.}
We follow the authors' video pipeline, which generates 25 frames at $576\times1024$ with SVD and takes no text.
The pipeline requires camera poses and depth, which the benchmark does not provide.
Depth is estimated for each frame with Depth-Anything-V2, as in the authors' pipeline.
The camera pose of each frame is estimated by RANSAC PnP from the $K$ conditioning points, using $T_s$ back-projected with the depth and the authors' fixed intrinsics as 3D points and $T_t$ as their 2D positions.
If fewer than four inliers remain, the pose is fitted to all conditioning points visible in that frame, and frames with fewer than four visible points take the pose of the nearest frame.
The authors' preprocessing then warps $V_s$ with these poses and adds the resulting camera displacement to the CoTracker trajectories of $V_s$; we also add the $K$ conditioning tracks, starting at $T_s$ in frame 0 and following $T_t$ afterwards.
Since this pipeline changes only the camera and keeps the scene motion of $V_s$, its dense trajectories conflict with the object motion requested in joint object--camera editing.

\paragraph{ReVideo.}
We follow the authors' single-region inference, which generates 14 frames with SVD and takes no text.
ReVideo takes the first frame of $V_s$, the frames of $V_s$ with an editing region masked out, and sparse trajectories; in place of the authors' trajectories, we give the $K$ conditioning tracks, starting at $T_s$ in frame 0 and following $T_t$ afterwards.
ReVideo regenerates the editing region and keeps the content of $V_s$ outside it, so a region smaller than the frame keeps the source camera.
We therefore use the whole frame as the region in the categories that change the camera.
In object-only editing, the region is the box enclosing the source and target positions of the conditioning points whose target track departs from the source track by more than 8 pixels in some frame, with a 32-pixel margin and a minimum size of $64\times64$; if no point qualifies, the whole frame is used.

\section{Additional Analyses of VLT}
\label{sec:supp_vlt_analysis}
\label{sec:supp_writing}
We analyze how track-input representation and individual reasoner inputs affect performance, and assess whether track-understanding training preserves the backbone's multimodal reasoning capability.

\subsection{Track-Input Representation Comparison}
\label{sec:vlt_interface}
\label{sec:supp_interface}
\label{sec:supp_interface_settings}
\begin{table}[t]
\centering\footnotesize
\setlength{\tabcolsep}{2pt}
\caption{\textbf{Comparison of track-input representations.}}
\label{tab:vlt_interface}
\begin{tabular}{@{}lcccc@{}}
\toprule
Track input & Pair$\uparrow$ & Read$\uparrow$ & Judge$\uparrow$ & Track gain$\uparrow$ \\
\midrule
\color{gray}\textit{Chance} & \color{gray}0.063 & \color{gray}0.113 & \color{gray}0.235 & \color{gray}0.000 \\
\midrule
Drawn on the frames & 0.307 & 0.686 & 0.532 & 0.257 \\
Learned track encoder & 0.000 & 0.363 & 0.367 & 0.044 \\
Coordinate text (\vlt, ours) & \textbf{0.528} & \textbf{0.831} & \textbf{0.612} & \textbf{0.320} \\
\bottomrule
\end{tabular}
\end{table}

We compare three ways of providing tracks to \vlt: coordinate text, rendered trajectories, and tokens from a learned track encoder.
Coordinate text represents tracks as coordinate sequences; rendered trajectories overlay colored paths, point IDs, and motion trails on the video frames; the learned encoder produces tokens indexed by track block, point, and time step.
All variants use the same eight-frame video samples, training data, adapters, loss, and number of training steps, and are evaluated on the same held-out questions.

As shown in \cref{tab:vlt_interface}, coordinate text achieves the highest scores on all four metrics.
It reaches 0.528 on Pair and 0.831 on Read, compared with 0.307 and 0.686 for rendered trajectories and 0.000 and 0.363 for the learned encoder.
Together with its higher Judge and Track gain scores, these results support coordinate text as the input representation for relating tracks to scene content and editing instructions.

\subsection{Input-Dependence Analysis}
\label{sec:vlt_cert}
We test how the source video, source tracks, target tracks, and editing instruction contribute to reasoner accuracy and generated trajectories.
Each intervention removes one block from the reasoner prompt without replacement, keeping the other inputs fixed.
For \vltgen, the generator retains its full track conditioning; only the reasoner prompt changes.

\begin{table}[t]
\centering\footnotesize
\setlength{\tabcolsep}{4pt}
\caption{\textbf{Input removal in \vlt.} Change in score when one input is removed from \vlt (c) of \cref{tab:vlt_curriculum}. Read and Judge average only the tasks that contain the removed input, so each row has its own task set. \sig\ paired bootstrap 95\% CI excludes zero. $^{\S}$Near zero by construction (see text). n/a: no task contains the input.}
\label{tab:supp_vlt_input_removal}
\begin{tabular}{@{}lccc@{}}
\toprule
Removed input & Pair & Read & Judge \\
\midrule
Source video & $-0.136$\sig & $-0.060$\sig & $-0.084$\sig \\
Source tracks & $-0.023$ & $-0.446$\sig & $-0.128$\sig \\
Target tracks & $-0.528^{\S}$ & n/a & $-0.157$\sig \\
Instruction & $-0.525^{\S}$ & n/a & $-0.290$\sig \\
\bottomrule
\end{tabular}
\end{table}

\paragraph{Reasoner.}
Removing any of the four inputs reduces at least one applicable \vlt score significantly (\cref{tab:supp_vlt_input_removal}).
Source-track removal has the largest effect on Read, reducing accuracy from 0.831 to 0.385.
For Judge, the mean decreases are 0.084, 0.128, 0.157, and 0.290 after removing the source video, source tracks, target tracks, and instruction, respectively; each mean covers the tasks that use that input.
Pair falls by 0.136 without the source video and by 0.023 without the source tracks, with only the former decrease significant.
Removing the target tracks or instruction reduces Pair to nearly zero: either removal makes matching and mismatching combinations difficult to distinguish by construction.
For the pretrained VLM and editing-only finetuning, significant decreases in the averaged scores are at most 0.054.
These interventions show that \vlt uses scene context, track geometry, and editing intent in answering the tasks.
Because the remaining prompt can still refer to a removed block, the measured effect includes both information loss and prompt mismatch.

\begin{table}[t]
\centering\footnotesize
\setlength{\tabcolsep}{2.6pt}
\caption{\textbf{Reasoner-input removal in \vltgen.}
Wan2.2-A14B, joint object--camera motion editing.
All conditions are evaluated on 100 pairs; the full-input reference is the result in \cref{tab:main}.
Parentheses show differences from the full-input reference.}
\label{tab:supp_gen_input_removal}
\resizebox{\columnwidth}{!}{\begin{tabular}{@{}lccc@{}}
\toprule
Removed from reasoner & EPE$_{\mathrm{c}}\downarrow$ & EPE$_{\mathrm{un}}\downarrow$ & PSNR$\uparrow$ \\
\midrule
None (\vltgen) & 6.45 & 8.71 & 16.47 \\
Source video & 10.19 ($+$3.74) & 13.81 ($+$5.10) & 15.35 ($-$1.12) \\
Source tracks & 9.82 ($+$3.37) & 13.62 ($+$4.91) & 15.38 ($-$1.09) \\
Target tracks & 11.33 ($+$4.88) & 15.15 ($+$6.44) & 15.49 ($-$0.98) \\
Instruction & 9.83 ($+$3.38) & 13.83 ($+$5.12) & 15.35 ($-$1.12) \\
\midrule
Coupling disabled & 11.90 ($+$5.45) & 15.66 ($+$6.95) & 15.00 ($-$1.47) \\
\bottomrule
\end{tabular}}
\end{table}

\paragraph{Generation.}
\Cref{tab:supp_gen_input_removal} compares the input-removal results with the full-input \vltgen result from \cref{tab:main}.
The full-input reference achieves EPE$_{\mathrm{c}}$ of 6.45\,px, EPE$_{\mathrm{un}}$ of 8.71\,px, and PSNR of 16.47\,dB.
All removal conditions have higher trajectory error and lower PSNR than this reference.
Among the four input-removal conditions, removing target tracks gives the highest EPE$_{\mathrm{c}}$ and EPE$_{\mathrm{un}}$ (11.33 and 15.15\,px).
Disabling the coupling gives errors of 11.90 and 15.66\,px and PSNR of 15.00\,dB.

\subsection{Backbone Capability Retention}
\label{sec:supp_general}
We evaluate whether track-understanding training preserves the multimodal reasoning capability of the Molmo2 backbone on MMStar.
Models (a)--(c) of \cref{tab:vlt_curriculum} are evaluated on its 1,500 questions, using accuracy as the metric.
We estimate 95\% confidence intervals for accuracy changes by paired bootstrap resampling of questions.

\begin{table}[t]
\centering\footnotesize
\setlength{\tabcolsep}{4pt}
\caption{\textbf{MMStar accuracy after finetuning.} Values in parentheses are the change from the pretrained VLM (a).}
\label{tab:supp_backbone}
\resizebox{\columnwidth}{!}{\begin{tabular}{@{}llc@{}}
\toprule
 & Model & MMStar$\uparrow$ \\
\midrule
(a) & VLM~\cite{molmo2} & 0.633 \\
(b) & (a) + editing-only finetuning & 0.622 ($-$0.011) \\
(c) & (a) + track-understanding tasks (\vlt) & 0.621 ($-$0.012) \\
\bottomrule
\end{tabular}}
\end{table}

Both finetuned models retain similar MMStar accuracy: 0.622 for editing-only finetuning and 0.621 for \vlt, compared with 0.633 for the pretrained VLM (\cref{tab:supp_backbone}).
Neither decrease is statistically significant, and the confidence intervals exclude drops larger than 3.1 percentage points.

\section{Token-Span Selection Analysis}
\label{sec:supp_token_spans}
\begin{figure*}[!t]
\centering
\includegraphics[width=\textwidth]{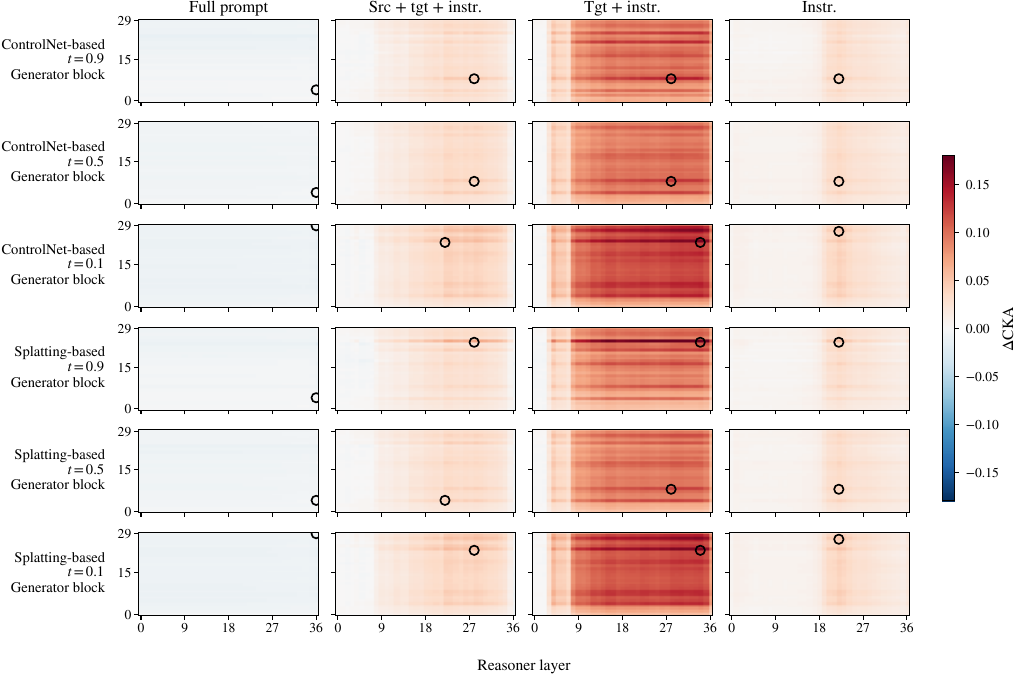}
\caption{\textbf{Comparison of prompt-token selections for the Wan2.1-1.3B generators.}
Each panel shows $\Delta\mathrm{CKA}$ (\cref{eq:delta_cka}) for every pair of reasoner layer and generator block.
We subtract source-video alignment from target-video alignment to reduce the contribution of shared scene content and assess information about the motion change; positive values indicate stronger target-video alignment.
Rows are the generator with a control branch (ControlNet-based) or latent splatting (Splatting-based) at normalized noise level $t$; columns are the token selections, which share one reasoner input and differ only in the pooled tokens.
Circles mark the maximum of each panel.}
\label{fig:token_spans_13b}
\end{figure*}
\begin{figure*}[!t]
\centering
\includegraphics[width=\textwidth]{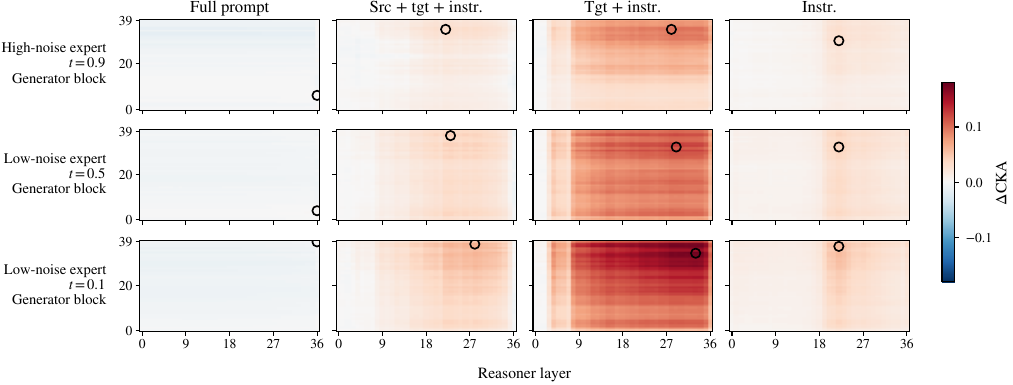}
\caption{\textbf{Comparison of prompt-token selections for the Wan2.2-A14B generator.}
Same analysis and color scale as \cref{fig:token_spans_13b}, with rows for the high- and low-noise experts at normalized noise level $t$.}
\label{fig:token_spans_a14b}
\end{figure*}

We examine which prompt-token representations capture the intended motion change beyond the scene content shared by the source and target videos.
The source video, source tracks, target tracks, and instruction are processed together in one reasoner forward pass.
We compare four token selections: the full prompt; source-track, target-track, and instruction spans; target-track and instruction spans; and the instruction span alone.
The instruction span includes the editing instruction and the following verification request.
These selections change which hidden states are pooled for analysis, while the reasoner input remains identical.

\paragraph{Paired representation comparison.}
We use linear centered kernel alignment (CKA)~\cite{cka} to compare the across-sample structure of reasoner and generator features.
Source and target videos share scene content but differ in motion.
A high CKA with target-video features may therefore reflect shared scene content rather than the intended motion change.
We subtract the alignment with source-video features to reduce this shared-scene contribution and assess how strongly the reasoner representations capture the motion change.
For each source--target video pair, we run the generator on noisy target and source latents using the same noise realization, noise level, and conditioning.
The target video is a reference for this analysis and is not supplied to \vlt.
Let $\bar H_r^{(l)}$ contain features mean-pooled over selected tokens $r$ at reasoner layer $l$, with one row per sample.
Let $G_{\mathrm{tgt}}^{(b)}$ and $G_{\mathrm{src}}^{(b)}$ contain the corresponding pooled generator features at block $b$.
We compute
\begin{equation}
\begin{aligned}
\Delta\mathrm{CKA}_{r,l,b}
={}&\mathrm{CKA}\!\left(\bar H_r^{(l)},G_{\mathrm{tgt}}^{(b)}\right)\\
&-\mathrm{CKA}\!\left(\bar H_r^{(l)},G_{\mathrm{src}}^{(b)}\right).
\end{aligned}
\label{eq:delta_cka}
\end{equation}
A positive $\Delta\mathrm{CKA}$ indicates stronger alignment with target-video features relative to the source-video reference.
We use this relative alignment to assess information about the motion change beyond the shared scene content.
Pooling is used only for this representation analysis; \vltgen transfers individual token states, whose effect on generation is evaluated separately in \cref{sec:exp_vltgen_coupling}.

\paragraph{Setup.}
We use 190 MoveBench editing pairs with recovered editing instructions and the final \vlt checkpoint with the eight-frame input used by \vltgen.
The generators are generator-only models trained without a reasoner: Wan2.1-1.3B with a control branch, Wan2.1-1.3B with latent splatting, and Wan2.2-A14B.
The A14B baseline uses a different track-input representation from \vltgen.
We extract generator features at normalized noise levels 0.9, 0.5, and 0.1, giving nine settings in total.
For A14B, the high-noise expert is used at 0.9 and the low-noise expert at 0.5 and 0.1.
We bootstrap editing pairs to obtain 95\% confidence intervals and shuffle reasoner--generator pairings to estimate a chance reference.

\paragraph{Results.}
The target-track and instruction spans have the highest $\Delta\mathrm{CKA}$ in all nine settings (\cref{fig:token_spans_13b,fig:token_spans_a14b}).
From reasoner layer 3 onward, they rank first at every layer--block pair.
After averaging over generator blocks, their peak $\Delta\mathrm{CKA}$ is 0.09--0.12 for the 1.3B generators and 0.06--0.13 for A14B.
At the peak reasoner layer, this is 2.8--3.0 times the next-best selection, with non-overlapping 95\% confidence intervals, and more than ten times the shuffled-pairing reference.
Adding source-track tokens lowers the peak to 0.02--0.05, close to the instruction-only selection; the full prompt remains near zero.

\paragraph{Implications for token selection.}
These results support transferring the target-track and instruction spans across the tested generator architectures, scales, and noise levels.
Their advantage holds across reasoner layers and generator blocks, allowing the span selection to remain fixed while layer routing is learned.
The lower scores obtained by adding source-track tokens or retaining only instruction tokens also support the chosen scope of transfer.
\FloatBarrier
\section{Coupling Design Analysis}
\label{sec:supp_decomp}
\begin{table*}[!t]
\centering\small
\setlength{\tabcolsep}{2.6pt}
\caption{\textbf{Ablation of token selection, layer routing, and value projection.}
Gray entries retain the corresponding design choice in Ours.
All variants use Wan2.1-1.3B as the generator backbone and are evaluated at 5K training iterations.
Generated-video metrics are the arithmetic means of per-sample scores over 200 source--target pairs, with 50 pairs randomly sampled from each of the four benchmark categories.
Input spec.\ is the pre-projection feature-energy fraction after removing the global mean and, for learnable queries, the fixed query-slot pattern.
Instr./track is instruction-sentence retention divided by target-track retention through the value projection (\cref{sec:supp_injected_metrics}).
Both diagnostics are dimensionless.
Learnable queries have one token type, leaving the projection-choice cells blank and Instr./track
undefined (n/a). The \onlygen row is the same track-conditioned generator trained without the reasoner, so it has no injected
values (n/a).}
\label{tab:vltgen_ablation}
\begin{tabular}{@{}llllccccccc@{}}
\toprule
 & \multicolumn{3}{c}{Coupling choice} & \multicolumn{2}{c}{Feature diagnostics} & \multicolumn{5}{c}{Generated video} \\
\cmidrule(lr){2-4}\cmidrule(lr){5-6}\cmidrule(lr){7-11}
 & Transferred tokens & Layer routing & Value projection & Input spec. & Instr./track & EPE$_{\text{c}}\downarrow$ & EPE$_{\text{un}}\downarrow$ & PSNR$\uparrow$ & SSIM$\uparrow$ & LPIPS$\downarrow$ \\
\midrule
\onlygen & \multicolumn{3}{c}{no coupling} & n/a & n/a & 29.04 & 31.11 & 11.77 & 0.493 & 0.575 \\
\midrule
\multicolumn{11}{@{}l}{\textit{Transferred from prompt states}} \\
Ours & track + instruction & learned, hard & per type & \textbf{0.419} & \textbf{0.667} & \textbf{14.36} & \textbf{15.56} & \textbf{12.87} & \textbf{0.523} & \textbf{0.504} \\
(1) & \color{gray}track + instruction & learned, soft & \color{gray}per type & 0.279 & 0.357 & 16.70 & 20.17 & 12.52 & 0.500 & 0.529 \\
(2) & \color{gray}track + instruction & fixed & \color{gray}per type & 0.417 & 0.598 & 18.61 & 22.40 & 12.41 & 0.492 & 0.552 \\
(3) & \color{gray}track + instruction & \color{gray}learned, hard & shared & 0.417 & 0.639 & 20.03 & 22.86 & 12.60 & 0.506 & 0.532 \\
\midrule
\multicolumn{11}{@{}l}{\textit{Transferred from learnable queries}} \\
(4) & learnable queries & \color{gray}learned, hard & & 0.013 & n/a & 21.57 & 25.57 & 12.53 & 0.508 & 0.537 \\
(5) & learnable queries & learned, soft & & 0.006 & n/a & 22.53 & 25.17 & 12.70 & 0.509 & 0.532 \\
(6) & learnable queries & fixed & & 0.014 & n/a & 20.78 & 23.71 & 12.63 & 0.499 & 0.536 \\
\bottomrule
\end{tabular}
\end{table*}

This section presents a coupling design ablation study (\cref{sec:exp_vltgen_coupling}) and an analysis of learned layer routing (\cref{sec:supp_routing}).
We first define the feature-energy and retention metrics in \cref{sec:supp_injected_metrics} to assess how coupling choices preserve feature variation across editing inputs and tokens through value projection, and how this retention differs between instruction and target-track tokens.

\subsection{Feature-Energy and Retention Metrics}
\label{sec:supp_injected_metrics}
\emph{Input spec.} assesses how much the features entering the projection vary across inputs and tokens, beyond shared fixed patterns.
This lets us compare the feature variation available after token selection and layer routing.
\emph{Retention} measures how much of the nonconstant feature-energy fraction is preserved through the value projection.
\emph{Instr./track} compares this retention between instruction and target-track tokens to assess whether the projection preserves instruction variation less strongly than track variation.

We use the same energy decomposition before and after the value projection.
Below, $v_{i,p}$ denotes the projected value for token $p$ of editing input $i$ at a given generator block; replacing it with the RMS-normalized reasoner state gives the pre-projection diagnostic.
We decompose it into a global mean $M$, an input-specific mean offset $A_i$, and a residual $E_{i,p}$:
\begin{equation}
\begin{aligned}
M &= \frac{\sum_i\sum_{p=1}^{P_i} v_{i,p}}{\sum_i P_i},\\
A_i &= \frac{1}{P_i}\sum_{p=1}^{P_i}v_{i,p}-M,\\
E_{i,p} &= v_{i,p}-M-A_i,
\end{aligned}
\end{equation}
where $P_i$ is the number of tokens in the group for input $i$.
Let $\mathcal E_M$, $\mathcal E_A$, and $\mathcal E_E$ denote the mean squared norms of these components, averaged over all input--token pairs.
The components are orthogonal under this token-weighted average, so their energies sum to the total value energy $\mathcal E_V$.
For prompt states, we compute
\begin{equation}
\mathrm{Spec.}=\frac{\mathcal E_A+\mathcal E_E}{\mathcal E_V}.
\end{equation}
This diagnostic includes both variation between input means and residual token variation within the variable-length prompt spans.

Learnable queries have fixed slots across inputs.
We additionally separate the input-invariant slot offset $B_p=\frac{1}{N}\sum_i v_{i,p}-M$, where $N$ is the number of editing inputs.
With $R_{i,p}=v_{i,p}-M-A_i-B_p$, their Spec. is $(\mathcal E_A+\mathcal E_R)/\mathcal E_V$, excluding both the global mean and the fixed slot pattern.
We aggregate across generator blocks by summing component energies before taking their ratio.
The table reports \emph{Input spec.}, obtained by applying this decomposition to the RMS-normalized states before projection and using all transferred tokens.
Its entries are fractions in $[0,1]$: for example, 0.419 corresponds to 41.9\% of the feature energy.
For variable-length prompt spans, this quantity includes residual variation across tokens as well as variation between input means.
It therefore describes feature variation, rather than a direct measure of semantic understanding.

\paragraph{Instruction--track balance and retention.}
We compute the energy fractions separately for target-track tokens and instruction tokens.
For token type $\tau$, let $s_{\tau}^{\mathrm{pre}}$ and $s_{\tau}^{\mathrm{post}}$ denote the fractions before and after value projection.
Retention is their ratio, $r_\tau=s_{\tau}^{\mathrm{post}}/s_{\tau}^{\mathrm{pre}}$.
The table reports the relative retention
\begin{equation}
\mathrm{Instr./track}
=\frac{r_{\mathrm{ins}}}{r_{\mathrm{trk}}}
=\frac{s_{\mathrm{ins}}^{\mathrm{post}}/s_{\mathrm{ins}}^{\mathrm{pre}}}
{s_{\mathrm{trk}}^{\mathrm{post}}/s_{\mathrm{trk}}^{\mathrm{pre}}}.
\end{equation}
A value below one means that the value projection retains a smaller proportion of the instruction energy fraction than of the track energy fraction.
Instruction diagnostics use the editing-instruction sentence separately from the fixed verification suffix.
We use the same blocks, excluding those that select the skip option, for pre- and post-projection retention.
The ratio is undefined for learnable queries, which have a single token type.

\paragraph{Diagnostic settings.}
The routing and value-projection diagnostics use 300 editing inputs.
Pre-projection fractions describe the RMS-normalized states entering the projection; retention measures how much of those fractions remains in the projected values.
We report individual retention values as percentages and their Instr./track ratio as a dimensionless number.

\subsection{Coupling Design Ablation Study}
\label{sec:exp_vltgen_coupling}
\input{fig/layer_routing/layer_routing_a14b.tex}
Using the diagnostics defined in \cref{sec:supp_injected_metrics}, we compare layer routing, value projection, and token selection at 5K training iterations (\cref{tab:vltgen_ablation}).
All variants use Wan2.1-1.3B as the generator backbone and are evaluated on the same 200 source--target pairs, randomly sampling 50 from each of the four categories in our benchmark.
For each generated-video metric, we report the arithmetic mean of the per-sample scores across all 200 pairs.
We examine the nonconstant feature fraction available to the coupling, its retention through projection, and the resulting video quality.
The full coupling achieves the lowest conditioned- and unconditioned-point errors and the best scores on all three target-video fidelity metrics among the tested variants.

\emph{Hard layer routing.}
Learned hard routing gives an Input spec.\ of 0.419, compared with 0.279 for soft layer mixtures.
In the separate target-track diagnostic, the pre-projection energy fractions are 38.7\% and 24.5\%, respectively.
Hard routing therefore preserves a larger nonconstant feature fraction before projection than soft layer mixtures.
Hard routing lowers EPE$_{\mathrm{c}}$ from 16.70 to 14.36\,px and EPE$_{\mathrm{un}}$ from 20.17 to 15.56\,px, while improving all three target-video fidelity metrics.
Compared with fixed layer mapping, learned hard routing reduces the two errors from 18.61 and 22.40\,px to 14.36 and 15.56\,px, respectively, and also improves fidelity.

\emph{Type-specific value projections.}
Under hard routing, type-specific value projections increase instruction-sentence retention from 11.7\% to 18.4\% compared with a shared value projection.
Target-track retention also increases, from 18.3\% to 27.6\%; Instr./track rises from 0.639 to 0.667, indicating better relative retention of instruction information.
This change also lowers EPE$_{\mathrm{c}}$ from 20.03 to 14.36\,px and EPE$_{\mathrm{un}}$ from 22.86 to 15.56\,px, while improving all three fidelity metrics.

\emph{Direct transfer of prompt states.}
With fixed layer mapping, prompt-state transfer gives an Input spec.\ of 0.417, compared with 0.014 for learnable queries (rows (2) and (6)).
Learnable queries remain nearly input-invariant under learned hard routing (0.013, row (4)).
With the same routing choice, direct prompt-state transfer lowers EPE$_{\mathrm{c}}$ from 21.57 to 14.36\,px and EPE$_{\mathrm{un}}$ from 25.57 to 15.56\,px, improvements of 7.21 and 10.01\,px, respectively.
It also improves all three fidelity metrics.
The tested prompt states thus retain more variation across edits and provide more effective conditioning than the tested learnable queries.

\subsection{Learned Layer Routing}
\label{sec:supp_routing}
Training changes the layer routing from its seeded initialization (\cref{fig:supp_routing}).
We summarize the logits by their softmax at temperature 0.5 without Gumbel noise and read each block's routing as its highest-logit candidate, which is the layer it reads at inference.
At initialization, every block gives a weight of 0.89 to its seeded candidate.
After training, 5 of the 40 blocks in the high-noise expert and 13 of the 40 blocks in the low-noise expert select a different layer, and all 18 changes are to a deeper reasoner layer (22, 27, or 32).
In these blocks, the weight of the seeded layer falls from 0.89 to a mean of 0.07 and 0.04, and to at most 0.25.
The changes occur in blocks 0--21, numbered from 0: blocks 0--1 of the high-noise expert and blocks 0--2 of the low-noise expert move from layer 7 to layer 32, and the low-noise expert moves blocks 8--16 from layers 12 and 17 to layers 27 and 32.
Blocks 22--39 keep their seeded layers in both experts.

\section{Additional Video Generation Results}
\label{sec:supp_generation}
This section presents additional video comparisons, a user study, an analysis of conditioning-track density, and a comparison with track-conditioned image-to-video methods.

\ifarxivversion
\subsection{Additional Video Comparisons}
\else
\subsection{Additional Quantitative Results}
\fi
\label{sec:supp_recon}
\label{sec:supp_epe_un}
\ifarxivversion
For additional video comparisons, see the Comparison section of our \href{https://minjung-s.github.io/thinkerbell}{project page}.
\else
For additional video comparisons, see the Comparison section of \texttt{ThinkerBell.html} included in the supplementary material.
\fi

\subsection{User Study}
\label{sec:supp_user_study}
\begin{figure*}[!t]
\centering
\captionsetup{skip=4pt}
\includegraphics[width=0.66\textwidth]{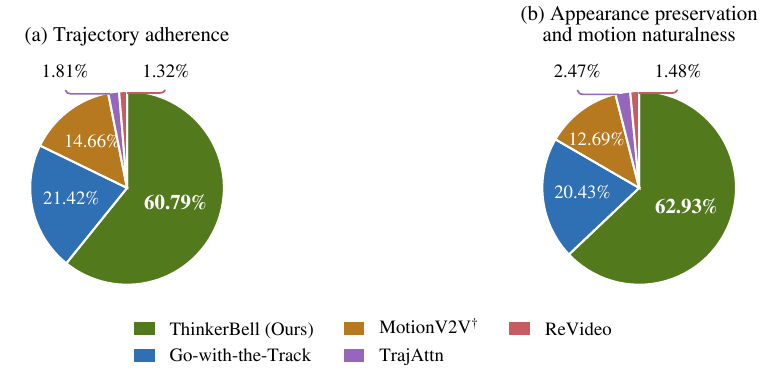}
\caption{\textbf{User preferences for video motion editing.}
The study involves 23 participants and 29 video samples randomly selected from the four benchmark categories.
Each chart shows the share of selections pooled across survey languages for (a) trajectory adherence and (b) appearance preservation and motion naturalness.
\vltgen receives a majority of selections under both criteria.
MotionV2V$^{\dagger}$ denotes our reimplementation.}
\label{fig:user_study}
\end{figure*}

\begin{figure*}[!t]
\centering
\captionsetup{skip=4pt}
\includegraphics[width=0.95\textwidth]{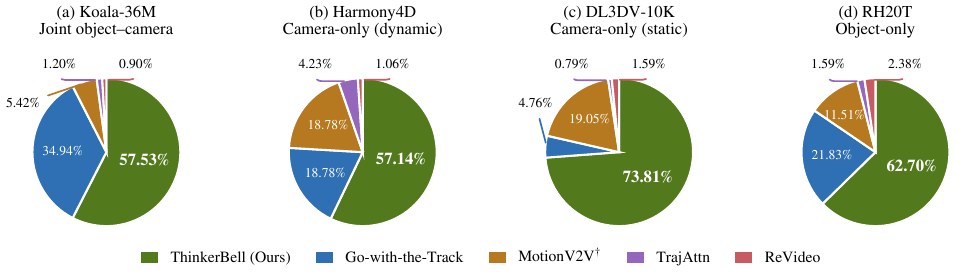}
\caption{\textbf{User preferences by benchmark category.}
Each chart pools selections from both evaluation questions and all survey languages within one category.
The percentages therefore summarize trajectory adherence and appearance preservation with motion naturalness jointly.
\vltgen receives a majority of selections in every category, with its highest share in camera-only editing of static scenes (DL3DV-10K).
Colors match \cref{fig:user_study}; MotionV2V$^{\dagger}$ denotes our reimplementation.}
\label{fig:user_study_categories}
\end{figure*}

\paragraph{Study setup.}
We conduct a user study with 23 participants on 29 video samples, randomly selecting 6--8 samples from each of the four benchmark categories in \cref{tab:main}.
The video samples are presented in randomized order.
For each sample, participants compare outputs from five methods: \vltgen, ReVideo, Go-with-the-Track, MotionV2V$^{\dagger}$, and TrajAttn.
They are shown the source video, target tracks, and a reference video illustrating the intended edit.
We use the benchmark target video as the reference, without identifying it to participants as the benchmark ground truth.
The reference video is not among the selectable method outputs.
The instructions identify the source video as the appearance reference for the objects and background, and the target tracks as the desired object or camera trajectories.
Participants are also told that a disappearing track point indicates occlusion of the corresponding scene part.

\begin{figure*}[!t]
\centering
\includegraphics[width=\textwidth]{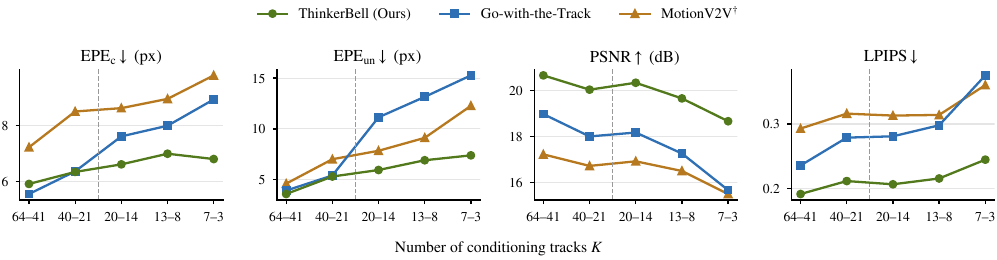}
\caption{\textbf{Effect of the number of conditioning tracks $K$.}
Bins are ordered from larger to smaller $K$.
The bins to the left of the dashed line use the same 300 pairs regenerated with more tracks ($K=41$--64: 170 pairs; $K=21$--40: 130 pairs).
The remaining bins partition the original results ($K=14$--20: 119 pairs; $K=8$--13: 90 pairs; $K=3$--7: 91 pairs).
EPE is measured at $480\times832$ resolution and averaged over evaluable pairs.
TrajAttn and ReVideo are omitted; their EPE$_{\mathrm{c}}$ is more than 15\,px higher than Ours in every bin.}
\label{fig:density}
\end{figure*}

\paragraph{Evaluation criteria.}
Participants select one preferred result for each of two questions.
The first evaluates \emph{trajectory adherence}: ``Which result video most accurately follows the specified motion trajectory?''
Participants use the target tracks and reference video to judge motion direction, pattern, and timing, setting aside visual quality and artifacts.
The second evaluates \emph{appearance preservation and motion naturalness}: ``Which result video best preserves the original appearance while generating natural-looking motion?''
Participants judge the colors and shapes of the objects and background relative to the source video, visual artifacts such as blurring, flickering, and distortion, and the naturalness of the motion.

\paragraph{Results.}
We aggregate all responses across survey languages and report the selection share for each method.
As shown in \cref{fig:user_study}, \vltgen receives 60.79\% of selections for trajectory adherence and 62.93\% for appearance preservation and motion naturalness, a majority under both criteria.
Go-with-the-Track is the next most preferred method, with 21.42\% and 20.43\%, respectively.
The preference margins of \vltgen are therefore 39.37 and 42.50 percentage points.
The agreement across the two criteria indicates that participants prefer \vltgen both for following the intended motion and for preserving scene appearance with natural-looking motion.

\paragraph{Preferences by benchmark category.}
\Cref{fig:user_study_categories} breaks down the preferences by benchmark category, pooling responses to both evaluation questions within each category.
\vltgen receives a majority of selections in all four categories: 57.53\% for joint object--camera editing (Koala-36M), 57.14\% for camera-only editing in dynamic scenes (Harmony4D), 73.81\% for camera-only editing in static scenes (DL3DV-10K), and 62.70\% for object-only editing (RH20T).
Its highest selection share is on DL3DV-10K, where MotionV2V$^{\dagger}$ is the next most preferred method at 19.05\%.
The closest comparison is on Koala-36M, where Go-with-the-Track receives 34.94\%, leaving a 22.59-percentage-point margin for \vltgen.
The preference for \vltgen thus extends across object motion, camera motion, and their joint editing.

\subsection{Effect of the Number of Conditioning Tracks}
\label{sec:supp_density}
We evaluate trajectory controllability and target-video fidelity as the number of conditioning tracks $K$ increases.
The analysis uses 300 pairs from RH20T, Harmony4D with different cameras, and DL3DV, covering object-only editing and camera-only editing in dynamic and static scenes.
We exclude Koala because its conditioning-track counts were selected manually.
We first group the randomly sampled $K=3$--20 results into three bins.
We then sample $K\sim\mathcal{U}\{21,\ldots,64\}$ for the same pairs, retain the original conditioning tracks within the enlarged set, and regenerate all methods with the same model checkpoints and generation settings.
Adding tracks can also change the subset of up to 12 points read by \vlt, the moving-to-static ratio in 26 object-only pairs with too few moving points, and the track-derived inputs of TrajAttn and ReVideo.
The comparison therefore measures the effect of increasing track density under this sampling procedure.

\Cref{fig:density} compares the absolute scores of \vltgen, Go-with-the-Track, and MotionV2V$^{\dagger}$ across the five bins.
\vltgen achieves the lowest EPE$_{\mathrm{c}}$ in the four bins covering $K=3$--40 and the lowest EPE$_{\mathrm{un}}$ in all five.
With $K=3$--7, its EPE$_{\mathrm{c}}$ is 6.79\,px versus 8.92\,px for Go-with-the-Track, and its EPE$_{\mathrm{un}}$ is 7.40\,px versus 15.26\,px.
The conditioned-point gap narrows to 0.02\,px at $K=21$--40 (6.33 versus 6.35\,px).
At $K=41$--64, Go-with-the-Track has lower EPE$_{\mathrm{c}}$ (5.54 versus 5.90\,px), while \vltgen retains lower EPE$_{\mathrm{un}}$ (3.60 versus 3.94\,px).
\vltgen also achieves the highest PSNR and lowest LPIPS in every bin, with a PSNR advantage of 1.68--2.99\,dB over Go-with-the-Track.
These results show the largest trajectory-error gaps between the two methods in the sparsest bin, while the fidelity advantage of \vltgen persists across the tested densities.
Each bin contains different video pairs, and EPE$_{\mathrm{un}}$ covers fewer points as more tracks are conditioned, so cross-bin trends reflect both track density and the evaluated samples.

\subsection{Comparison with Track-Conditioned I2V Methods}
\label{sec:supp_i2v}
We compare \vltgen with ATI~\cite{ati} and Wan-Move~\cite{wanmove}, which generate a video from a single reference frame and target tracks.
Our main benchmark does not provide a shared first frame, so evaluating I2V methods requires a separate construction.
Following MotionV2V~\cite{motionv2v}, we split a video at frame $t_0$: the target contains frames $t_0$ through $t_0+80$, and the source contains frames $t_0$ through $t_0-80$ in reverse chronological order.
Both videos therefore start from the same frame.
We construct 50 joint object--camera pairs from Koala-36M using the same tracking and annotation pipeline, excluding near-static pairs whose tracks differ by less than 20\,px on average between the two halves.
ATI and Wan-Move receive the shared first frame, target tracks, and target description.
\vltgen receives the source video, source and target tracks, target description, and editing instruction.
All methods use the same conditioning tracks, with the $K$ sampling and metrics of \cref{tab:main}.

\begin{table}[t]
\centering\footnotesize
\setlength{\tabcolsep}{3pt}
\caption{\textbf{Comparison with track-conditioned I2V methods} on 50 joint object--camera pairs that share their first frame. EPE excludes one pair without a ground-truth point visible in its last frame.}
\label{tab:supp_i2v}
\begin{tabular}{@{}lccccc@{}}
\toprule
Method & EPE$_{\mathrm{c}}\downarrow$ & EPE$_{\mathrm{un}}\downarrow$ & PSNR$\uparrow$ & SSIM$\uparrow$ & LPIPS$\downarrow$ \\
\midrule
ATI~\cite{ati} & 9.37 & 13.79 & \textbf{17.30} & 0.627 & 0.310 \\
Wan-Move~\cite{wanmove} & 10.93 & 15.66 & 16.29 & 0.611 & 0.321 \\
\midrule
\vltgen (Ours) & \textbf{7.02} & \textbf{11.33} & 17.22 & \textbf{0.634} & \textbf{0.286} \\
\bottomrule
\end{tabular}
\end{table}

\vltgen achieves EPE$_{\mathrm{c}}$ of 7.02\,px, compared with 9.37\,px for ATI and 10.93\,px for Wan-Move (\cref{tab:supp_i2v}).
Its EPE$_{\mathrm{un}}$ is also lower: 11.33\,px versus 13.79 and 15.66\,px, respectively.
PSNR is similar to ATI (17.22 versus 17.30\,dB) and higher than Wan-Move (16.29\,dB); LPIPS is lower than both.
These results compare the methods under their respective input formats: the I2V methods continue from the reference frame, while \vltgen regenerates it from the source video.
Reversing the source half can also make irreversible actions appear unnatural in the input to \vltgen.

\section{Applications}
\label{sec:supp_applications}
We apply \vltgen to video stabilization, driving scene simulation, and camera reshoot without task-specific training.
Each application specifies the desired motion through target tracks and an editing instruction.
\ifarxivversion
The qualitative examples, including edits with user hand-drawn track inputs, are available on our \href{https://minjung-s.github.io/thinkerbell}{project page}.
\else
The qualitative examples, including edits with user hand-drawn track inputs, are available on the project page, \texttt{ThinkerBell.html}.
\fi

\paragraph{Video stabilization.}
We stabilize handheld video by specifying target tracks that follow a smooth camera path.
We demonstrate this application on the project page using clips from DeepStab, with target tracks extracted from the paired stabilized videos.

\paragraph{Driving scene simulation.}
We edit recorded driving videos to change the camera viewpoint or driving path.
We demonstrate these edits on the project page using driving sequences from Argoverse~2.
For side-view synthesis, we use a front-camera video as the source and specify a left- or right-side viewpoint with tracks matched to the recorded side-camera video.
The two recordings can be offset by up to 5\,s.
For driving-path changes, we project static LiDAR points along a camera path that shifts from the recorded ego pose by one lane (3.5\,m) to the left or right over the clip, with the camera turning toward the new lane.
These tracks control the static scene; other vehicles are not re-simulated for the new path.

\paragraph{Camera reshoot.}
We change camera motion after filming by constructing target tracks along a new camera path.
We demonstrate this application on the project page using shots from Tears of Steel.
We lift user-selected points to 3D using depth and camera poses estimated by VGGT, then reproject them along an orbit, crane, or dolly path.

These selected examples illustrate the applications qualitatively. The driving-path and camera-reshoot examples lack ground-truth target videos, and the side-view recordings are not temporally aligned references.

\section{Computational Cost}
\label{sec:supp_cost}
The computational cost of \vltgen is dominated by its Wan2.2-A14B generator backbone.
For the closest comparison, both \vltgen and MotionV2V$^{\dagger}$ use this backbone to generate 81 frames at $480\times832$ with 50 denoising steps on one H200 GPU.
Their recorded median generation times are 983 and 963\,s, respectively, with median peak allocated GPU memory of 136.9 and 126.3\,GiB.
The differences are approximately 20\,s (2\%) and 10.6\,GiB (8.4\%).
Thus, most of the runtime and memory footprint is already present in the same-backbone MotionV2V$^{\dagger}$ baseline, with comparatively modest overhead from our reasoning and coupling components.
\vlt runs once per edit in approximately 0.4\,s, and its features are reused across all denoising steps; the reported generation time uses cached \vlt features.

For context, Go-with-the-Track, TrajAttn, and ReVideo take median times of 291, 294, and 25\,s under their respective inference configurations, generating 49, 25, and 14 frames.
TrajAttn and ReVideo record median peak reserved GPU memory of 38.4 and 17.2\,GiB; GPU memory was not recorded for Go-with-the-Track.
These measurements come from Koala-36M runs on one H200 GPU, with 12 inputs for \vltgen and 105 for each baseline.
Cross-method costs reflect differences in generator backbones, output lengths, resolutions, and inference procedures; the SVD-based implementations report reserved memory, while \vltgen and MotionV2V$^{\dagger}$ report allocated memory.

\section{Discussion}
\label{sec:supp_discussion}
Effective coupling requires reasoner representations that vary with the intended edit and retain that variation when transferred to the generator.
The learnable-query interfaces we test, following~\cite{metaquery,instructx}, have nearly input-invariant features, whereas the selected prompt states retain variation across edits (\cref{sec:exp_vltgen_coupling}).
This comparison motivates evaluating the content transferred through a coupling alongside its effect on generated videos.
Our design selects edit-relevant prompt spans, routes reasoner layers to generator blocks, and adapts the two token types with separate value projections.
The generation ablations link these choices to trajectory controllability and target-video fidelity.

\section{Limitations}
\label{sec:supp_limitations}
\begin{figure*}[t]
\centering\footnotesize
\newcommand{\rflab}[1]{\parbox{0.085\textwidth}{\centering #1}}
\setlength{\tabcolsep}{1.5pt}
\begin{tabular}{cc@{\hspace{8pt}}c}
 & (a) Failure of \vltgen & (b) Failure of Go-with-the-Track \\
\rflab{Source} & \raisebox{-.5\height}{\includegraphics[width=0.44\textwidth]{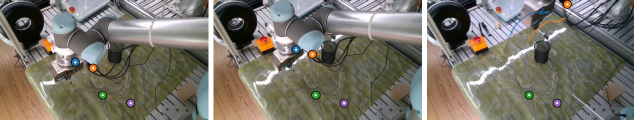}} & \raisebox{-.5\height}{\includegraphics[width=0.44\textwidth]{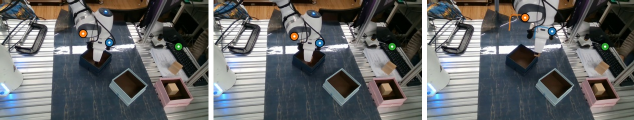}} \\[1pt]
\rflab{GT} & \raisebox{-.5\height}{\includegraphics[width=0.44\textwidth]{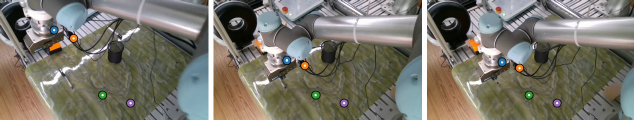}} & \raisebox{-.5\height}{\includegraphics[width=0.44\textwidth]{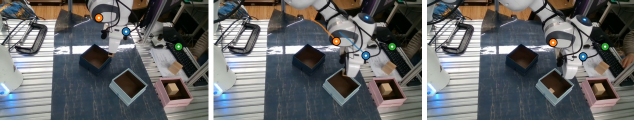}} \\[1pt]
\rflab{Go-with-\\the-Track} & \raisebox{-.5\height}{\includegraphics[width=0.44\textwidth]{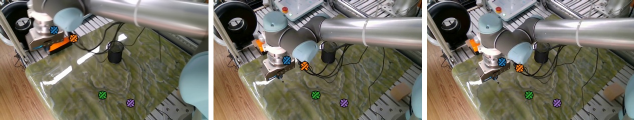}} & \raisebox{-.5\height}{\includegraphics[width=0.44\textwidth]{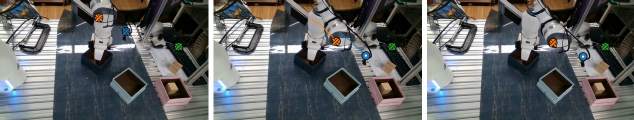}} \\[1pt]
\rflab{\vltgen} & \raisebox{-.5\height}{\includegraphics[width=0.44\textwidth]{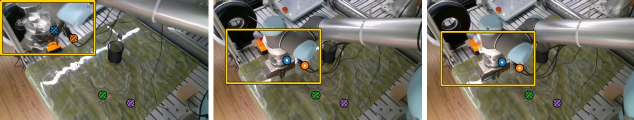}} & \raisebox{-.5\height}{\includegraphics[width=0.44\textwidth]{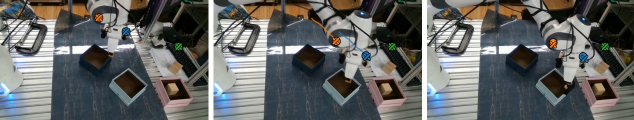}} \\[1pt]
\rflab{\vltgen\\(zoom)} & \raisebox{-.5\height}{\includegraphics[width=0.44\textwidth]{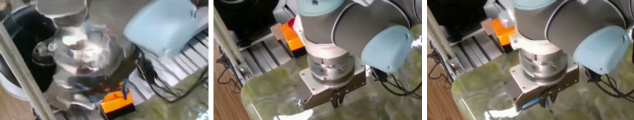}} & \\
\end{tabular}
\caption{\textbf{Failure cases on RH20T.}
Columns show frames 0, 40, and 80.
Filled dots mark the target tracks, and crosses of the same color mark the points re-tracked for EPE; no cross is drawn when the tracker reports the point as invisible, which EPE penalizes as a visibility mismatch when the target point is visible.
(a) A failure of \vltgen: it generates a blurry first frame; the bottom row enlarges the boxed gripper region.
The gripper becomes sharp within the first few frames and follows the target tracks, but the tracker, starting from the blurry frame, loses the gripper points.
(b) A failure of Go-with-the-Track: the gripper head detaches and moves independently of the robot arm, illustrating ambiguity in whether sparse conditioning points refer to an individual part or the whole articulated object.
The gripper deviates from its target track, so the elevated EPE reflects a failure of trajectory controllability in the generated motion.}
\label{fig:supp_rh20t_fail}
\end{figure*}

\vltgen sometimes produces blur or duplicated structures in the initial frames.
Because trajectory evaluation re-tracks the generated video from the first frame, these artifacts can make the queried points difficult to localize and cause the tracker to lose them.
Tracking failures can persist even after the generated object becomes sharp and follows the target tracks, as illustrated in \cref{fig:supp_rh20t_fail}a.
When the tracker reports a point as invisible in the generated video but visible in the target video, our evaluation protocol applies a 52-pixel visibility-mismatch penalty (\cref{sec:supp_benchmarks}).
Consequently, early-frame artifacts can affect trajectory scores across subsequent frames, even after visual quality improves.
This failure mode shows how early-frame appearance artifacts can affect both visual quality and trajectory evaluation, motivating more consistent appearance from the start of generation.

Our 2D trajectory conditioning is easy for users to specify and supports object and camera motion editing, including object rotations guided by track visibility and editing instructions.
However, camera motions involving substantial 3D viewpoint changes, such as orbit-view edits, remain challenging to specify using sparse 2D trajectories alone.
For these edits, 3D trajectories or camera-parameter conditioning could provide more direct control, although users may find such target conditions harder to construct.

\begingroup
    \small

\endgroup
\end{bibunit}
\end{document}